\documentclass[prodmode]{acmsmall} 

\usepackage{verbatim}
\usepackage{color}
 \usepackage{cite}
 \usepackage{slashbox}
\usepackage{hyperref}
\usepackage{graphicx}
\usepackage{graphics,epstopdf}
\usepackage{array}
\usepackage{float}
\usepackage{amsmath,amssymb,stmaryrd}
\usepackage[tight]{subfigure}
\newcommand{\dset}[1]{\textsf{#1}}
\newcommand{\CM}{\mathbf{C}}

\newcommand{\bx}{\mathbf{x}}

\newcommand{\bracketof}[1]{\left[ #1 \right]}

\newcommand{\bool}[1]{\left\llbracket #1 \right\rrbracket}

\newcommand{\cost}{\mathbf{c}}
\newcommand{\costof}[2][]{\cost_{#1}\!\bracketof{#2}}

\newcommand{\tcost}{\bar{\cost}}

\newcommand{\calD}{\mathcal{D}}
\newcommand{\calE}{\mathcal{E}}

\newcommand{\calS}{\mathcal{S}}

\newcommand{\bbR}{\mathbb{R}}

\newcommand{\expt}{\mathop{\calE}\limits}

\usepackage{url}
\usepackage{amsmath}
\usepackage{stmaryrd}
\usepackage{algorithm}
\usepackage{verbatim}
\usepackage{caption}

\usepackage{todonotes}

\runningfoot{}

\begin{document}

\markboth{T. Jan et al.}{Soft Methodology for 
Cost-and-error Sensitive Classification}

\title{Soft Methodology for 
Cost-and-error Sensitive Classification}
\author{Te-Kang Jan
\affil{Institute of Information Science, Academia Sinica}
Da-Wei Wang
\affil{Institute of Information Science, Academia Sinica}
Chi-Hung Lin
\affil{National Yang-Ming University}
Hsuan-Tien Lin
\affil{National Taiwan University}
}

\begin{abstract}
Many real-world data mining applications need varying cost for different types of classification errors and thus call for cost-sensitive classification algorithms. Existing algorithms for  cost-sensitive classification are successful in terms of minimizing the cost, but can result in a high error rate as the trade-off. The high error rate holds back the practical use of those algorithms. In this paper, we propose a novel cost-sensitive classification methodology that takes both the cost and the error rate into account. The methodology, called soft cost-sensitive classification, is established from a multicriteria optimization problem of the cost and the error rate, and can be viewed as regularizing cost-sensitive classification with the error rate. The simple methodology allows immediate improvements of existing cost-sensitive classification algorithms. Experiments on the benchmark and the real-world data sets show that our proposed methodology indeed achieves lower test error rates and similar (sometimes lower) test costs than existing cost-sensitive classification algorithms. We also demonstrate that the methodology can be extended for considering the weighted error 
rate instead of the original error rate. This extension is useful for tackling unbalanced classification problems.

\end{abstract}

%




%

\maketitle
\vfill\eject 
\section{Introduction}
Classification is important for machine learning and data mining~\cite{han2011data,hall2009weka}. 
Traditionally, the regular classification problem aims at minimizing the rate of misclassification errors. In many real-world applications, however, different types of errors are often charged with different costs. 
For instance, in bacteria classification, mis-classifying a Gram-positive species as a Gram-negative one leads to totally ineffective treatments and
is hence more serious than mis-classifying a Gram-positive species as another Gram-positive one~\cite{bibm2011,schleifer2009classification}. 
Similar application needs are shared by targeted marketing, information retrieval, medical decision making, object recognition and intrusion detection~\cite{freitas2011building,lee2002toward,tan1993cost,fan2000multiple,sun2007cost,NAb04}, and can be formalized as the cost-sensitive classification problem. 
In fact, cost-sensitive classification can be used to express any finite-choice and bounded-loss supervised learning
problems~\cite{ABey05}. Thus, it has been attracting
much research attention in recent years, in terms of both new algorithms and new applications~\cite{filter,tan1993cost,thesis1,bernstein2005toward,thesis2,thesis3,bibm2011}.

Studies in cost-sensitive classification often reveal a trade-off between cost and error rate \cite{thesis2,thesis1,thesis3}.
Mature regular classification algorithms can
achieve significantly lower error rate than their cost-sensitive counterparts, but result in higher expected cost; state-of-the-art cost-sensitive
classification algorithms can reach significantly lower expected cost than their regular classification counterparts, but are often
at the expense of higher error rate.
In addition, cost-sensitive classification algorithms are ``sensitive'' to large cost components
and can thus be conservative or even ``paranoid'' in order to avoid making any big mistakes. 
The sensitivity
makes cost-sensitive classification algorithms prone to overfitting the data or the cost. In fact,
it has been observed that for some simpler classification tasks, cost-sensitive
classification algorithms are inferior to regular classification ones in terms of even the expected test cost because of the overfitting~\cite{thesis1,thesis2}. 

The expense of high error rate and the potential risk of overfitting 
holds back the practical use of cost-sensitive classification algorithms. Arguably, applications call for classifiers that can reach low cost \textit{and} low error rate.
The problem of obtaining such a classifier has been studied only for binary cost-sensitive classification \cite{soft1} and 
strategy for decision tree \cite{Webb1996}, but the more general problem for multiclass cost-sensitive classification is yet to be tackled.

In this paper, we propose a methodology to tackle the problem. The methodology takes both the cost and the error rate into account and matches the realistic needs better.
We name the methodology \textit{soft cost-sensitive classification}
to distinguish it from existing \textit{hard cost-sensitive classification}
algorithms that focus on only the cost. The methodology is designed by formulating
the associated problem as a multicriteria optimization task \cite{moobook}: one criterion
being the cost and the other being the error rate.
Then, the methodology solves the task by the weighted sum approach for multicriteria optimization \cite{1105511}.
The simplicity of the weighted sum approach allows immediate reuse of modern cost-sensitive
classification algorithms as the core tool. In other words, with our proposed methodology,
promising (hard) cost-sensitive classification algorithms can be immediately improved via soft cost-sensitive classification, with performance guarantees on cost and error rate supported
by the theory behind multicriteria optimization.
 
Error rate, however, is sometimes not the basic criterion of interest. For instance, many cost-sensitive classification data sets in the real world are also unbalanced, such as the  intrusion detection data set in KDD Cup 1999~\cite{bay2000uci}. For such an unbalanced data set, the error rate favors only the majority classes and is thus less meaningful in assessing the quality of classification results. Then, the weighted error rate that balances the influence of each class can be more meaningful. We extend the proposed methodology to consider the weighted error rate instead of the error rate. The extended methodology can then be used to improve the performance of cost-sensitive classification algorithms for unbalanced classification problems.

We conduct a  complete  comparison  to validate the performance of the proposed methodology. The comparison involves not only twenty-two benchmark and two real-world data sets, but also uses four state-of-the-art (hard) cost-sensitive classification algorithms as well as their soft siblings.
To the best of our knowledge, the comparison is the most extensive empirical study on multiclass cost-sensitive classification in terms of the numbers of data sets and algorithms. Experimental results suggest that soft cost-sensitive classification can indeed achieve both low cost and low error rate. In particular, soft cost-sensitive classification algorithms out-perform regular ones in terms of the test cost on most of the data sets. In addition, soft cost-sensitive classification algorithms reach significantly lower test error rate than their hard siblings, while achieving similar (sometimes better) test cost. The observations are consistent across three different sets of tasks: the traditional benchmark tasks in cost-sensitive classification~\cite{PD99}, new benchmark tasks designed for examining the effect of using large cost components, and the real-world medical task for classifying bacteria~\cite{bibm2011}. 

We also conduct experiments on unbalanced classification tasks for validating the extended methodology. The unbalanced data sets include not only the benchmark data sets but also a real-world task, the KDD 1999 data set on intrusion detection \cite{bay2000uci}. The results justify that soft cost-sensitive classification can consider cost and weighted error rate jointly to reach better performance.


The paper is organized as follows. We formally introduce the regular and the cost-sensitive classification problems in Section \ref{section2}, and discuss related works on cost-sensitive classification. Then, we present the proposed
methodology of soft cost-sensitive classification in Section \ref{section3}. We discuss the empirical performance of the proposed methodology on the  benchmark and the real-world data sets in Section
\ref{section4}. Finally, we conclude in Section~\ref{section5}.

A short version of the paper appeared in 18th ACM SIGKDD Conference on Knowledge Discovery and Data Mining\cite{soft2012}. The paper is then enriched by
\begin{enumerate}
\item introducing another state-of-the-art cost-sensitive classification algorithm \cite{ZZho06} in Section~\ref{section2}, and including it in the experimental comparison in Section~\ref{section4} with two types of different costs that are added for making a fair comparison with this algorithm;
\item extending the proposed methodology to take  both weighted error rate and cost into account in Section~\ref{section3}, and validate its performance in Section~\ref{section4};    
\item studying the issue of parameter selection for soft cost-sensitive classification substantially in Section~\ref{section4}.
\end{enumerate}


\section{Cost-sensitive Classification}
\label{section2}

We shall start by defining the regular classification problem and then extend it to the cost-sensitive one. Then, we briefly
review existing works on cost-sensitive classification.

In the regular classification problem, we are given a training set
$\calS = \{ (\bx_n,y_n) \}_{n=1}^N$, where the input vector $\bx_n$
belongs to some domain $\cal{X} \subseteq $ $\mathbb{R}^D$, the label $y_n$
comes from the set $\cal{Y}$ $ = \{1,\dots,K\}$ and each example
$(\bx_n,y_n)$ is drawn independently from an unknown distribution $\calD$
on $\cal{X} \times \cal{Y}$. The task of regular classification is to use the
training set $\calS$ to find a classifier $g\colon
{\cal{X}}\rightarrow{\cal{Y}}$ such that the expected error rate
 $E(g) = \expt_{(\bx, y) \sim \calD} \bool{y  \neq  g(\bx) }$ is small,\footnote{The Boolean operation $\bool{\cdot}$ is $1$ when the
  argument is true and$~0$ otherwise.} where the expected error rate $E(g)$ penalizes every type
of mis-classification error equally.

Cost-sensitive classification extends regular classification
by charging different cost for different types of classification errors.
We adopt the example-dependent setting of cost-sensitive classification, which
is rather general and can be used to express other popular settings~\cite{JLan05c,filter,thesis1,thesis2,thesis3}.
The example-dependent setting couples each example $(\bx, y)$ with a
cost vector$~\cost \in \left[ 0, \infty \right)^K$, where the $k$-th component of $\cost$ quantifies the cost 
for predicting the example $\bx$ as class $k$. 
The cost $\cost[y]$ of the intended \linebreak class $y$ is naturally assumed to be $0$, the minimum cost.
Consider a cost-sensitive training set$~\calS_c=\{(\bx_n,y_{n},\cost_{n})\}_{n=1}^{N}$, where each cost-sensitive training example $(\bx_n,y_{n},\cost_{n})$ is drawn independently
from an unknown cost-sensitive distribution $\calD_c$ on$~{\cal{X} \times \cal{Y} }\times \left[ 0 , \infty \right)^K$,
the task of cost-sensitive classification is to use 
$\calS_c$ to find a classifier $g\colon
{\cal{X}}\rightarrow{\cal{Y}}$ such that the expected cost
$
 E_c (g) = \expt_{(\bx, y, \cost) \sim \calD_c} \costof{g(\bx)} 
$
is small.

One special case of the example-dependent setting is the class-dependent setting, in which the cost
vectors $\cost$ are taken from the $y$-th row of a cost matrix$~\CM\colon {\cal{Y} \times \cal{Y}} \rightarrow \left[ 0, \infty \right)^K$. Each entry $\CM(y,k)$ of the cost matrix represents
the cost for predicting a \linebreak class-$y$ example as class $k$. The special case is commonly used in some applications and some benchmark experiments~\cite{bibm2011,thesis1,thesis3}.

Regular classification can be viewed as a special case of the class-dependent setting, which is in term
a special case of the example-dependent setting. In particular, take a cost matrix that contains$~0$ in the diagonals and $1$ elsewhere, which equivalently corresponds to the regular cost vectors $\tcost_y$ with entries$~\tcost_y[k]~=~\bool{y \neq k}$. Then, the expected cost$~E_c(g)$ with respect to $\{\tcost_y\}$ is the same as the expected error rate$~E(g)$.
In other words, regular classification algorithms 
can be viewed as ``wiping out'' the given cost information and replacing it with a na{\"i}ve cost matrix.
Intuitively, such algorithms may not work well for cost-sensitive classification
because of the wiping out. 

Another special case of the class-dependent setting considers a cost matrix where row $y$ equals $w_y\cdot \tcost_y$, with some weight$~w_y \geq 0$ for each $y.$ The weights can be used to adjust the influence of each class, and are widely used when solving unbalanced classification problems. This special case is commonly named weighted classification.

%
%
%
%

Existing cost-sensitive classification algorithms can be grouped to two categories: the binary ($K=2$) cases and the multiclass~($K>2$) cases. Binary cost-sensitive classification is well-understood in theory and in practice. In particular, every binary cost-sensitive classification problem can be reduced to a binary regular classification one by 
re-weighting the examples based on the cost\cite{elkan2001foundations,BZ03}. Multiclass cost-sensitive classification, however, is more difficult than the binary one, and is an ongoing research topic.

MetaCost\cite{PD99} is one of the earliest multiclass cost-sensitive classification algorithms and it can only be applied to the class-dependent setting. MetaCost makes any regular classification algorithm cost-sensitive by re-labeling the training examples. Somehow the re-labeling procedure depends on an overly-ideal assumption, which makes it hard to rigorously analyze the performance of MetaCost in theory. Many other early approaches suffer from similar shortcomings~\cite{DD1}.

In order to design multiclass cost-sensitive classification algorithms with stronger theoretical guarantees, modern cost-sensitive classification algorithms are mostly reduction-based, which allows not only reusing mature existing algorithms for cost-sensitive classification, but also extending existing theoretical results to the area of cost-sensitive classification. For instance, \cite{NAb04} reduces the multiclass cost-sensitive classification problem into several multiclass weighted classification problems using a boosting-style method and some intermediate traditional classifiers. The reduction is somehow too sophisticated for practical use. 

Zhou and Liu proposed another reduction approach (CSZL; \cite{ZZho06}) from multiclass cost-sensitive classification to multiclass weighted classification based on re-weighting with the solution to a linear system. The CSZL approach can only work in the class-dependent setting. When the cost matrix is consistent (i.e. coefficient matrix of the linear system is not of full rank), CSZL comes with sound theoretical guarantees for choosing the the weights, and then plugs these weights into some weighted classification algorithm as an internal learner; otherwise, CSZL decomposes the multiclass cost-sensitive classification problem into several binary cost-sensitive classification problems based on pairwise comparisons of the classes to get an approximate solution \cite{ZZho06}. 

There are quite a few other studies on reducing multiclass cost-sensitive classification to binary cost-sensitive classification by decomposing the multiclass problem with a suitable structure and embedding the cost vectors into the weights in those binary classification problems. For instance, cost-sensitive one-versus-one (CSOVO; \cite{thesis1}) and  weighted all-pair (WAP; \cite{ABey05}) are also based on pairwise comparisons of the classes. Another leading approach within the family is cost-sensitive filter tree (CSFT; \cite{filter}), which is based on a single-elimination tournament of competing classes. 

Yet another family of approaches reduce the multiclass cost-sensitive classification problem into regression ones by embedding the cost vectors in the real-valued labels instead of the weights \cite{thesis4}. A promising representative of the family is to reduce to one-sided regression (OSR; \cite{thesis2}). 

Based on some earlier comparisons on  general benchmark data sets \cite{thesis3,thesis2}, OSR, CSOVO and CSFT are some of the leading algorithms that can reach state-of-the-art performance.  Each algorithm corresponds to a popular sibling for regular classification. In particular, the common one-versus-all decomposition (OVA) \cite{299108} is the special case of OSR, the one-versus-one decomposition (OVO) \cite{299108} is the special case of CSOVO, and the modern filter tree decomposition (FT) \cite{filter} is the special case of CSFT. The regular classification algorithms, OVA, OVO and FT, do not consider any cost during their training. On the other hand, the cost-sensitive ones, OSR, CSOVO and CSFT, respect the cost faithfully during their training.  

Note that the regular classification sibling for CSZL is not as explicit as the other cost-sensitive classification algorithms. When the cost matrix consists of $\{ \tcost_y \}$, the cost is consistent 
for CSZL and its corresponding linear system can be solved by setting all classes to be of equal  weights. Thus, the regular classification sibling of CSZL is the regular classification sibling of its internal learner. Because CSZL takes one-versus-one decomposition for the inconsistent cost, we consider (weighted) OVO as the internal learner for CSZL for the consistent cost in this work. Hence the regular classification sibling of CSZL is simply OVO.

%

\begin{figure}[t]
  \centering
  \includegraphics[width=.49\columnwidth]{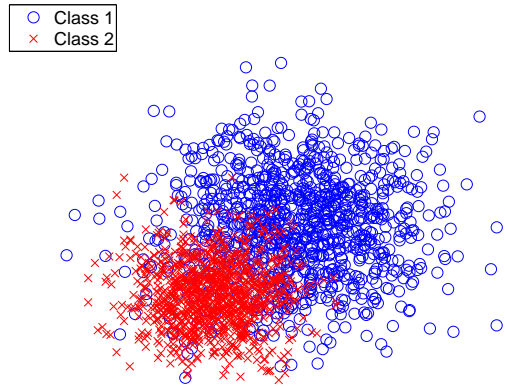}\\
  \caption{a two-dimensional artificial data set}
  \label{figartificial}
\end{figure}
%

\section{Soft Cost-sensitive Classification}
\label{section3}

\begin{figure}[t]
   \centering
   \includegraphics[width=.49\columnwidth]{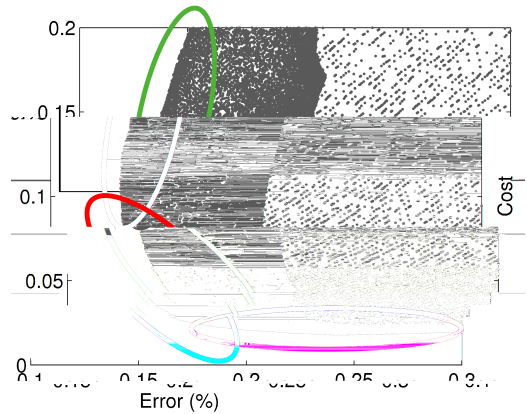}\\
  \caption{the different goals of regular (green), cost-sensitive (red) and soft cost-sensitive (blue) classification algorithms}
  \label{figsoft}
 \end{figure}

The difference between regular and cost-sensitive classification is illustrated
with a binary and two-dimensional artificial data set shown in Figure~\ref{figartificial}. Class$~1$ is generated from a Gaussian distribution of standard deviation
$\frac{4}{5}$; class $2$ is generated from a Gaussian distribution of standard deviation $\frac{1}{2}$; the centers of the two classes are of $\sqrt{2}$ apart. We consider a cost matrix of $\left[ \begin{array}{cc} 0 & 1\\ 30 & 0 \end{array} \right]$. Then, we
enumerate many linear classifiers in $\bbR^2$ and evaluate their average error and average cost. The results are plotted
in Figure~\ref{figsoft}. Each black point represents the achieved (\textsf{error}, \textsf{cost}) of one linear classifier.%
\footnote{Ideally, the points should be dense. The uncrowded part comes from simulating with a finite enumeration process.} 
We can see that there is a region of low-cost linear classifiers, as circled in red. There is also a region of low-error linear classifiers,
as circled in green. Modern cost-sensitive classification algorithms are designed to seek for \textit{something} in the red region, which contains classifiers with a wide range of different errors. Traditional regular classification algorithms, on the other hand,
are designed to locate something in the green region (without using the cost information), which is far from the lowest achievable cost.
In other words, there is a trade-off between the cost and the error, while cost-sensitive and regular classification each
takes the trade-off to the extreme.

Many real-world applications, however, do not need the extreme classifiers in the red and green regions, but call for classifiers with both low cost \textit{and} low error rate as depicted in the blue region in Figure~\ref{figsoft}. In particular, the applications take the cost to be the subjective
measure of performance and the error to be the objective safety-check as the basic criterion. The blue
region improves the green one (regular) by taking the cost into account;
the blue region also improves the red one (cost-sensitive) by keeping
the error under control. The three regions, as depicted, are not meant to be disjoint. 
The blue region may contain the better cost-sensitive classifiers in its intersection with the green region,
and the better regular classifiers in its intersection with the red region.
%

Figure~\ref{figsoft} results from a simple artificial data set for the illustrative purpose. When applying more sophisticated classifiers on real-world data sets, the set of achievable (\textsf{error}, \textsf{cost}) may be of a more complicated shape---possibly non-convex, for instance. Somehow the essence of
the problem remains the same: cost-sensitive classification only knocks down the cost and results in a red region at the bottom; regular classification only considers the error and lands on a green region at the left; our proposed methodology focuses on a blue region at the left-bottom, hopefully achieving the better for both criteria.

Formally speaking, regular classification algorithm is a process from $\mathcal{S}$ to $g$ such that $E(g)$ is small. Cost-sensitive classification
algorithm, on the other hand, is a process from$\mbox{~}\mathcal{S}_c$ to $g$ such that $E_c(g)$ is small. We now want a process from $\mathcal{S}_c$ to $g$
such that both $E(g)$ and $E_c(g)$ are small, which can be written as
\begin{eqnarray}
\min_g \mathbf{E}(g) = [E_c(g), E(g)] \mbox{ subject to all feasible } g. \label{moo_first}
\end{eqnarray}
The vector $\mathbf{E}$ represents the two criteria of interest. 

Such a problem belongs
to multicriteria optimization~\cite{moobook}, which deals with multiple objective functions.
The general form of multicriteria optimization is
\begin{eqnarray}
\label{moo}
\min_{g}  \mathbf{F}(g)=[F_1(g), F_2(g),\dots, F_M(g) ] \nonumber \\ \mbox{ subject to all feasible } g,
\end{eqnarray}
where $M$ is the number of criteria.
For a multicriteria optimization problem (\ref{moo}), often there is no global optimal solution $g^*$ that is the best in terms
of every dimension (criterion) within $\mathbf{F}$. Instead, the goal \linebreak of (\ref{moo}) is to seek for the set of ``better'' solutions, usually referred to as the Pareto-optimal front~\cite{350037}. Formally speaking, consider two feasible candidates $g_1$ \linebreak and $g_2$.
The candidate$~g_1$ is said to \textit{dominate} $g_2$ \linebreak if $F_m(g_1) \leq F_m(g_2)$ for all $m$ while $F_i(g_1) < F_i(g_2)$ for some~$i$. The Pareto-optimal front is the set of all non-dominated solutions~\cite{moobook}. 

Solving the multicriteria	optimization problem is not an
easy task, and there are many sophisticated techniques, including evolutionary algorithms like Non-dominated Sorting Genetic Algorithms~\cite{deb2002fast} and Strength Pareto Evolutionary Algorithms~\cite{corne2001pesa}. One important
family of techniques is to transform the problem to a single-criterion optimization one that we
are more familiar with. A simple yet popular approach of the family considers
a non-negative linear combination of all the criteria $F_m$, which is called the weighted sum approach~\cite{1105511}.
In particular, the weighted sum approach solves the following optimization problem:
\begin{eqnarray}
\label{moo2}
\min_g \sum_{m=1}^M \alpha_m F_m(g)   \mbox{ subject to all feasible } g,
\end{eqnarray}
where $\alpha_m \ge 0$ is the weight (importance) of the $m$-th criterion. By varying the values of $\alpha_m$, the weighted
sum approach identifies \textit{some} of the solutions that are on the tangential of the Pareto-optimal front~\cite{moobook}. 
The drawback of the approach~\cite{moodrawback} is that \textit{not all} the solutions within the Pareto-optimal front can be found
when the achievable set of $\mathbf{F}(g)$ is non-convex.

We can reach the goal of getting a low-cost and low-error classifier by formulating a multicriteria optimization problem with $M = 2$, $F_1(g) = E_c(g)$ \linebreak and $F_2(g) = E(g)$.
Without loss of generality, \linebreak let $\alpha_1 = 1 - \alpha$ and $\alpha_2 = \alpha$ for $\alpha \in [0, 1]$, the weighted sum approach solves
\begin{eqnarray}
\label{softeq0}
\min_{g}  (1-\alpha)E_c(g) + \alpha E(g),
\end{eqnarray}
which is the same as
\begin{eqnarray}
\label{softeq}
\min_{g}  \expt_{(\bx, y, \cost) \sim \calD_c}  (1-\alpha)\Bigl(\cost[g(\bx)]\Bigr) + \alpha\Bigl(\tcost_{y}[g(\bx)]\Bigr)
\end{eqnarray}
with the regular cost vectors $\tcost_y$ defined in Section~\ref{section2}. For any given $\alpha$,
such an optimization problem is exactly a cost-sensitive classification one with modified cost vectors $\tilde{\cost} = (1 - \alpha)\cost + \alpha \tcost_y$. 
Then, modern cost-sensitive classification algorithms can be applied to locate a decent $g$, which
would belong to the Pareto-optimal front with respect to $E_c(g)$ and~$E(g)$.

The weighted sum approach has also been implicitly taken by other algorithms in machine learning. For instance, \cite{sculley2010combined} combines the pairwise ranking criterion and squared regression criterion and shows that the resulting algorithm achieves the best performance on both criteria. Our proposed methodology similarly utilizes the simplicity of the weighted sum approach to allow seamless reuse of modern cost-sensitive classification algorithms. If other techniques for multicriteria optimization (such as evolutionary computation) are taken instead, new algorithms need to be designed to accompany the techniques. Given the prevalence of promising cost-sensitive classification algorithms (see Section \ref{section2}), we thus choose to study only the weighted sum approach.

The parameter $\alpha$ in (\ref{softeq0}) can be intuitively explained as a soft control of the trade-off between cost and error, with $\alpha = 0$ and $\alpha = 1$
being the two extremes.
The traditional (hard) cost-sensitive classification problem is a special case of soft cost-sensitive classification with $\alpha = 0$. On
the other hand, the regular classification problem is a special case of soft cost-sensitive classification with $\alpha = 1$. 

Another explanation behind (\ref{softeq0}) is regularization. 
From Figure \ref{figsoft}, there are many low-cost classifiers in the red region.
When picking one classifier using only the limited information in the training set $\mathcal{S}_c$,
the classifier can be over-fitting. The added \linebreak term $\alpha E(g)$ can be viewed
as restricting the number of low-cost classifiers by only favoring those with lower error rate. This similar explanation can be found \linebreak from \cite{soft1}, which considers cost-sensitive classification in
the binary case. Furthermore, the restriction is similar to common regularization schemes, where
a penalty term on complexity is used to limit the number of candidate classifiers~\cite{lfd}.

\begin{figure}[t]
            \centering
   \includegraphics[scale=0.31]{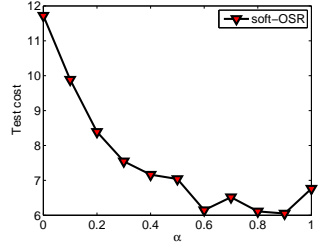}\\
         \caption{the effect of the regularization parameter $\mbox{~}\bm{\alpha}$ on soft cost-sensitive classification}
          \label{figb}
          \end{figure}

We illustrate the regularization property of soft cost-sensitive classification
with the data set \dset{vowel} as an example. The details of the experimental procedures will be introduced 
in Section~\ref{section4}. The test cost of soft cost-sensitive classification with various $\alpha$ when coupled with the one-sided regression (OSR) algorithm is shown in Figure \ref{figb}. For this data set, the lowest test cost does not happen at $\alpha = 0$ (hard cost-sensitive) nor $\alpha = 1$ (non cost-sensitive). By choosing the regularization parameter $\alpha$ appropriately, some intermediate, non-zero values of $\alpha$ (soft cost-sensitive) could lead to better test performance. The figure reveals the potential of soft cost-sensitive classification not only to improve the test error with the added $\alpha E(g)$ term
during optimization, but also to possibly improve the test cost with the effect of regularization.

The simplicity of (\ref{softeq0}) allows soft cost-sensitive classification to modify the basic criterion easily. For instance,
in an unbalanced classification problem, the weighted error rate $E_w(g) = \expt_{(\bx, y) \sim \calD} w_y  \cdot  \tcost_y [g(\bx)] $ instead of $E(g)$ is often used to respect the influence of each class properly. If we replace $E(g)$ with $E_w(g)$ in (\ref{softeq0}), we get 
\begin{eqnarray}
\label{softeq111}
\min_{g}  \expt_{(\bx, y, \cost) \sim \calD_c}  (1-\alpha)\Bigl(\cost[g(\bx)]\Bigr) + \alpha \Bigl(w_y \cdot   \tcost_{y}[g(\bx)]\Bigr)
\end{eqnarray}
The modified methodology (\ref{softeq111}) can also be solved by modern cost-sensitive classification algorithms to get a decent $g$ for both $E_c$ and $E_w$.

\section{Experiments}
\label{section4}

In this section, we set up experiments to validate the usefulness of the proposed methodology of soft cost-sensitive classification in various procedures. We take four state-of-the-art multiclass cost-sensitive classification algorithms (see Section 2). Then we examine if the proposed methodology can improve them. The four algorithms are one-sided regression (OSR), cost-sensitive one-versus-one (CSOVO), cost-sensitive filter tree (CSFT) and cost-sensitive classification  by Zhou and Liu (CSZL). We also include their regular classification siblings, one-versus-all (OVA), one-versus-one (OVO), and filter tree (FT) for comparisons. Note that OVO is also the regular classification sibling of CSZL and hence is denoted as OVO/ZL. 


We couple all the algorithms with the support vector machine (SVM) \cite{VV98} with the perceptron  kernel \cite{HL08a} as the internal learner for the reduced problem, and take LIBSVM \cite{CC01a} as the SVM solver.\footnote{We use the cost-sensitive SVM implementation at \url{http://www.csie.ntu.edu.tw/~htlin/program/cssvm/}} The regularization parameter $\lambda$ of SVM is chosen within $\{2^{10}, 2^{7}, \dots, 2^{-2}\}$. For the hard cost-sensitive classification algorithms, the best parameter setting is chosen by minimizing the 5-fold cross-validation cost. For the regular classification algorithms, which are not supposed to access any cost information in training or in validation, the best parameter $\lambda$ is chosen by minimizing the 5-fold cross-validation error.	We will study more about selecting the parameter $\alpha$ for soft cost-sensitive classification in  Section \ref{subsection41}.
%

%


We consider four sets of tasks: the traditional benchmark tasks for balancing the influence of each class, a real-world biomedical task for classifying bacteria (see Section 1), new benchmark tasks for emphasizing some of the classes, and the KDD Cup 1999 task for intrusion detection. These four tasks will demonstrate that soft cost-sensitive classification is useful both as a general algorithmic methodology and as a specific application tool.

\subsection{Parameter Selection for Soft Cost-Sensitive Classification}
\label{subsection41}

An important issue for soft cost-sensitive classification is to choose  the regularization parameter $\alpha$ properly. In particular, given two criteria of interest in soft cost-sensitive classification, it is non-trivial to decide the cross-validation criterion for picking the best parameter combination. We study two possible
scenarios: For the first one, we simply take the cost to be the cross-validation criterion, with ties broken by choosing the largest $\alpha$ (most regularization); for the second one, we intend to choose a parameter that leads to both low error and low cost, and hence  use $\max$(\textsf{error}, \textsf{normalized~cost})  as the cross-validation criterion to be minimized. We report the results by running OSR on eight data sets:  \dset{iris}, \dset{wine}, \dset{glass}, 
\dset{vehicle}, \dset{vowel}, \dset{segment}, \dset{dna}, \dset{satimage}, while similar observations have been found on other datasets and algorithms. For the cost, we take the benchmark one which will be introduced in  Section \ref{subsec1}. We normalize the sum of the cost matrix to be equal to sum of the na{\"i}ve cost matrix that  contains $\{ \tcost_y \}$.

The results are shown in  Table \ref{tbl:eq} and Table \ref{tbl:eq2} using a pairwise one-tailed $t$-test of significance  level $0.1$. The results confirm the trade-off between error and cost. In particular, CV by cost reaches lower cost  than CV by $\max$(\textsf{error}, \textsf{normalized~cost}) in 3 out  of 8 data sets, but CV by $\max$(\textsf{error}, \textsf{normalized~cost}) achieves lower error rate in 6 out of 8 data sets. Based on the study, we decide to use CV by cost for its simplicity and its better performance on the major criterion (cost).

 \begin{table}[h]
\caption{average test cost results for two validation criteria, with $t$-test for cost}
 \label{tbl:eq}
\begin{center}
\begin{tabular}{l|c|c|c}
   &  & CV by   &  \\
      & CV by cost  & $\max$(\textsf{error}, \textsf{normalized~cost})  & $t$-test \\
\hline
\dset{iris}  & $18.78\pm3.71$ & $21.58 \pm4.37$ & $\thickapprox$ \\
\hline
\dset{wine}  & $ 12.28\pm2.96 $ &  $ 11.39 \pm2.70$ & $\thickapprox$ \\
\hline
\dset{glass}  & $129.42\pm9.50$ & $ 139.72\pm9.32$ & $\bigcirc$ \\
\hline
\dset{vehicle}  &  $ 95.43\pm10.41$ & $109.65\pm9.07$  & $\bigcirc$ \\
\hline
\dset{vowel} & $ 6.42\pm1.10$ & $ 7.04 \pm1.03$ & $\thickapprox$ \\
\hline
\dset{segment} & $13.02\pm1.08 $ & $13.33\pm1.02$& $\thickapprox$\\
\hline
\dset{dna} & $  22.76\pm1.46 $ & $ 23.23\pm1.26$ & $\thickapprox$ \\
\hline	
\dset{satimage} & $34.86\pm2.13$ & $ 37.56\pm1.93$ & $\bigcirc$ \\
\hline
\end{tabular}

\end{center}
  \scalebox{0.85}{
  \begin{tabular}{c@{$\colon$}l}
  $\bigcirc$ & CV by cost significantly better than the other procedure\\ 
    $\times$ & CV by cost significantly worse than the other procedure\\ 
  $\thickapprox$ & otherwise
  \end{tabular}
  }
\end{table}

 \begin{table}[h]
\caption{average test error rate results for two validation criteria, with $t$-test for error rate}
 \label{tbl:eq2}
\begin{center}
\begin{tabular}{l|r|c|c}
   &  & CV by   &  \\
      & CV by cost  & $\max$(\textsf{error}, \textsf{normalized~cost})  & $t$-test \\
\hline
\dset{iris}  & $4.73\pm0.73$ & $ 5.00\pm0.71$ & $\thickapprox$ \\
\hline
\dset{wine}  & $ 2.44\pm0.38$ & $ 1.88\pm0.42$ & $\times$ \\
\hline
\dset{glass}  & $31.94\pm1.21$ & $ 31.11\pm0.98$ & $\thickapprox$ \\
\hline
\dset{vehicle}  & $22.78\pm0.72$ & $  21.56\pm0.77$ & $\times$ \\
\hline
\dset{vowel} &   $ 2.01\pm0.34$ &$ 1.59\pm0.22$  & $\times$ \\
\hline
\dset{segment} & $2.96\pm0.17$ &  $2.71\pm0.13 $ &  $\times$\\
\hline
\dset{dna} & $  4.87\pm0.27 $ & $ 4.16\pm0.14$ & $\times$ \\
\hline	
\dset{satimage} & $9.01\pm0.33$ & $ 7.30\pm0.13$ & $\times$ \\
\hline
\end{tabular}

\end{center}
  \scalebox{0.85}{
  \begin{tabular}{c@{$\colon$}l}
  $\bigcirc$ & CV by cost significantly better than the other procedure\\ 
    $\times$ & CV by cost significantly worse than the other procedure\\ 
  $\thickapprox$ & otherwise
  \end{tabular}
  }
\end{table}

%
%

\subsection{Comparison on Benchmark Tasks }
\label{sub42}

Twenty-two real-world data sets (\dset{iris}, \dset{wine}, \dset{glass}, 
\dset{vehicle}, \dset{vowel}, \dset{segment}, \dset{dna}, \dset{satimage}, 
\dset{usps}, \dset{zoo}, \dset{yeast}, \dset{pageblock}, \dset{anneal}, \dset{solar}, \dset{splice}, \dset{ecoli}, \dset{nursery}, \dset{soybean}, \dset{arrhythmia, optdigits, mfeat, pendigit}) are
used in our next experiments. All data sets come from the UCI Machine Learning 
Repository~\cite{SHe98} except \dset{usps}~\cite{JH94}. In each run of the experiment, we randomly separate each data set with $75\%$ of the examples for 
training and the rest $25\%$ for testing. All the input vectors in the training set are linearly scaled to $[0, 1]$ and then the input vectors in the test set are scaled accordingly. These data sets do not contain any cost information and we generate two types of costs for each benchmark data set, one is inconsistent cost, and another is consistent cost (see Section \ref{section2}). 

%

\subsubsection{Inconsistant Cost Matrix}\label{subsec1}

We first generate costs similar to the procedure used by \cite{ABey05,thesis1,thesis2}. In particular, the benchmark is class-dependent and is based
 on a cost  matrix $\CM(y,k)$, where the diagonal entries $\CM(y,y)$  are 0, and the other entries $\CM(y,k)$ are uniformly sampled  from $\left[0, \frac{|\{n:y_n=k\}|}{|\{n:y_n=y\}|} \right]$. This means that mis-classifying a rare class as a frequent one is of a high cost in expectation. We further scale every $\CM(y, k)$  to $\left[0,1\right]$ by dividing it with the largest component in $\CM$. We then record
the average test cost and their standard errors for all algorithms
over 20 random runs in Table~\ref{tbl:softresult}. We also report the average test errors in Table~\ref{tbl:softresult2}.



\begin{table*}[h]
  \caption{average test cost ($\cdot 10^{-3}$) on benchmark data sets for inconsistant cost \protect\linebreak (note that the regular sibling of CSZL is also OVO)} 
  \label{tbl:softresult}
  {
  \scalebox{0.48}{
  \begin{tabular}{
      c|
      r@{$\pm$}l@{\hspace{7pt}}
      r@{$\pm$}l@{\hspace{7pt}}
      r@{$\pm$}l@{\hspace{7pt}}
      r@{$\pm$}l@{\hspace{7pt}}
      r@{$\pm$}l@{\hspace{7pt}}
       r@{$\pm$}l@{\hspace{7pt}}
        r@{$\pm$}l@{\hspace{7pt}}
            r@{$\pm$}l@{\hspace{7pt}}
        r@{$\pm$}l@{\hspace{7pt}}
              r@{$\pm$}l@{\hspace{7pt}}
        r@{$\pm$}l@{\hspace{7pt}}
    }
    data set & 
    \multicolumn{2}{c}{OVA} &
    \multicolumn{2}{c}{OSR} &
    \multicolumn{2}{c}{soft-OSR} &
      \multicolumn{2}{c}{FT} &
    \multicolumn{2}{c}{CSFT} &     
    \multicolumn{2}{c}{soft-CSFT}&
    \multicolumn{2}{c}{OVO/ZL} &
    \multicolumn{2}{c}{CSOVO} &   
    \multicolumn{2}{c}{soft-CSOVO} &  
    \multicolumn{2}{c}{CSZL} &     
    \multicolumn{2}{c}{soft-CSZL}\\
    \hline
    \dset{iris} & $\mathbf{18.34}$ & $\mathbf{4.48}$ & $\mathbf{17.21}$ & $\mathbf{3.84}$ & $\mathbf{18.79}$ & $\mathbf{3.72}$ & $23.80$ & $5.21$ & $19.54$ & $4.67$ & $\mathbf{15.91}$ & $\mathbf{3.55}^*$ & $21.93$ & $4.99$ & $20.74$ & $4.32$ & $\mathbf{19.34}$ & $\mathbf{4.26}$ & $20.56$ & $3.88$ & $21.20$ & $3.89$ \\
\dset{wine} & $\mathbf{12.98}$ & $\mathbf{3.37}$ & $\mathbf{13.42}$ & $\mathbf{2.55}$ & $\mathbf{12.97}$ & $\mathbf{2.93}$ & $15.21$ & $3.49$ & $\mathbf{11.87}$ & $\mathbf{3.09}$ & $15.62$ & $4.44$ & $15.04$ & $4.05$ & $\mathbf{11.45}$ & $\mathbf{3.53}^*$ & $\mathbf{13.91}$ & $\mathbf{4.33}$ & $\mathbf{13.71}$ & $\mathbf{3.71}$ & $16.26$ & $4.11$ \\
\dset{glass} & $159.19$ & $10.37$ & $\mathbf{126.84}$ & $\mathbf{9.71}^*$ & $\mathbf{129.42}$ & $\mathbf{9.51}$ & $151.06$ & $10.20$ & $143.78$ & $8.66$ & $143.22$ & $9.85$ & $145.90$ & $10.36$ & $\mathbf{128.56}$ & $\mathbf{9.77}$ & $\mathbf{132.69}$ & $\mathbf{9.62}$ & $\mathbf{136.44}$ & $\mathbf{9.56}$ & $141.11$ & $10.71$ \\
\dset{vehicle} & $114.14$ & $9.08$ & $\mathbf{95.33}$ & $\mathbf{10.29}^*$ & $\mathbf{97.81}$ & $\mathbf{10.85}$ & $112.48$ & $7.71$ & $\mathbf{105.58}$ & $\mathbf{10.90}$ & $106.74$ & $11.27$ & $112.31$ & $8.82$ & $\mathbf{103.63}$ & $\mathbf{11.17}$ & $\mathbf{97.34}$ & $\mathbf{11.16}$ & $\mathbf{100.69}$ & $\mathbf{10.72}$ & $\mathbf{98.23}$ & $\mathbf{11.05}$ \\
\dset{vowel} & $\mathbf{6.76}$ & $\mathbf{0.93}$ & $11.72$ & $1.44$ & $\mathbf{6.43}$ & $\mathbf{1.11}$ & $9.53$ & $1.31$ & $13.71$ & $1.58$ & $11.87$ & $1.47$ & $\mathbf{6.29}$ & $\mathbf{0.94}$ & $9.58$ & $1.08$ & $\mathbf{6.82}$ & $\mathbf{0.90}$ & $\mathbf{6.21}$ & $\mathbf{0.96}^*$ & $\mathbf{6.38}$ & $\mathbf{0.95}$ \\
\dset{segment} & $\mathbf{14.02}$ & $\mathbf{1.17}$ & $\mathbf{13.84}$ & $\mathbf{0.94}$ & $\mathbf{13.03}$ & $\mathbf{1.08}^*$ & $15.01$ & $1.33$ & $14.17$ & $1.15$ & $15.36$ & $1.26$ & $14.15$ & $1.18$ & $\mathbf{14.00}$ & $\mathbf{1.11}$ & $\mathbf{14.10}$ & $\mathbf{1.31}$ & $\mathbf{13.95}$ & $\mathbf{1.21}$ & $14.39$ & $1.27$ \\
\dset{dna} & $24.43$ & $1.26$ & $24.40$ & $1.55$ & $\mathbf{22.76}$ & $\mathbf{1.47}^*$ & $27.94$ & $2.34$ & $31.49$ & $2.09$ & $29.23$ & $2.28$ & $24.51$ & $1.37$ & $28.26$ & $2.04$ & $24.51$ & $1.52$ & $25.04$ & $1.41$ & $\mathbf{23.46}$ & $\mathbf{1.32}$ \\
\dset{satimage} & $40.20$ & $2.08$ & $\mathbf{35.04}$ & $\mathbf{2.16}$ & $\mathbf{34.86}$ & $\mathbf{2.11}^*$ & $41.98$ & $2.08$ & $40.16$ & $2.10$ & $39.63$ & $2.23$ & $40.43$ & $1.92$ & $\mathbf{36.49}$ & $\mathbf{2.27}$ & $\mathbf{36.46}$ & $\mathbf{2.31}$ & $38.70$ & $2.04$ & $38.97$ & $1.89$ \\
\dset{usps} & $6.87$ & $0.28$ & $7.32$ & $0.23$ & $\mathbf{6.58}$ & $\mathbf{0.27}^*$ & $9.05$ & $0.29$ & $8.97$ & $0.40$ & $8.59$ & $0.27$ & $7.08$ & $0.27$ & $7.20$ & $0.26$ & $6.98$ & $0.25$ & $7.12$ & $0.24$ & $7.08$ & $0.25$ \\
\dset{zoo} & $6.26$ & $1.81$ & $\mathbf{2.49}$ & $\mathbf{0.50}$ & $6.02$ & $1.84$ & $5.32$ & $1.54$ & $3.55$ & $0.91$ & $3.87$ & $1.30$ & $6.62$ & $1.85$ & $2.77$ & $0.64$ & $\mathbf{2.26}$ & $\mathbf{0.47}^*$ & $5.75$ & $1.81$ & $6.29$ & $1.78$ \\
\dset{yeast} & $36.66$ & $3.37$ & $\mathbf{0.58}$ & $\mathbf{0.07}$ & $\mathbf{0.58}$ & $\mathbf{0.07}$ & $38.97$ & $3.88$ & $\mathbf{0.62}$ & $\mathbf{0.09}$ & $0.64$ & $0.09$ & $39.71$ & $3.62$ & $\mathbf{0.55}$ & $\mathbf{0.08}^*$ & $\mathbf{0.55}$ & $\mathbf{0.08}$ & $0.66$ & $0.11$ & $0.64$ & $0.09$ \\
\dset{pageblock} & $2.80$ & $0.48$ & $0.18$ & $0.04$ & $0.19$ & $0.04$ & $2.78$ & $0.48$ & $\mathbf{0.16}$ & $\mathbf{0.03}$ & $\mathbf{0.16}$ & $\mathbf{0.03}^*$ & $2.59$ & $0.45$ & $\mathbf{0.16}$ & $\mathbf{0.03}$ & $\mathbf{0.16}$ & $\mathbf{0.03}$ & $\mathbf{0.16}$ & $\mathbf{0.03}$ & $\mathbf{0.16}$ & $\mathbf{0.03}$ \\
\dset{anneal} & $0.85$ & $0.23$ & $\mathbf{0.35}$ & $\mathbf{0.12}^*$ & $\mathbf{0.38}$ & $\mathbf{0.13}$ & $0.85$ & $0.23$ & $0.58$ & $0.16$ & $0.64$ & $0.16$ & $0.83$ & $0.23$ & $0.61$ & $0.16$ & $0.67$ & $0.17$ & $0.61$ & $0.15$ & $0.67$ & $0.16$ \\
\dset{solar} & $46.08$ & $6.53$ & $25.35$ & $4.06$ & $25.32$ & $4.05$ & $47.18$ & $7.14$ & $20.54$ & $2.64$ & $20.43$ & $2.06$ & $44.51$ & $6.31$ & $\mathbf{18.04}$ & $\mathbf{1.94}$ & $\mathbf{17.89}$ & $\mathbf{1.95}^*$ & $22.08$ & $2.46$ & $21.16$ & $2.49$ \\
\dset{splice} & $14.01$ & $0.84$ & $\mathbf{12.59}$ & $\mathbf{1.11}$ & $\mathbf{12.85}$ & $\mathbf{0.71}$ & $16.64$ & $0.79$ & $18.19$ & $1.62$ & $16.06$ & $1.17$ & $13.97$ & $0.76$ & $17.06$ & $1.26$ & $13.28$ & $0.88$ & $\mathbf{12.39}$ & $\mathbf{0.82}$ & $\mathbf{12.27}$ & $\mathbf{0.77}^*$ \\
\dset{ecoli} & $17.11$ & $2.85$ & $1.27$ & $0.31$ & $0.92$ & $0.18$ & $20.43$ & $4.49$ & $\mathbf{0.85}$ & $\mathbf{0.14}$ & $1.96$ & $1.13$ & $19.93$ & $2.61$ & $1.35$ & $0.49$ & $1.11$ & $0.41$ & $\mathbf{0.76}$ & $\mathbf{0.12}^*$ & $0.94$ & $0.15$ \\
\dset{nursery} & $0.62$ & $0.20$ & $\mathbf{0.00}$ & $\mathbf{0.00}$ & $\mathbf{0.00}$ & $\mathbf{0.00}^*$ & $1.42$ & $0.45$ & $0.00$ & $0.00$ & $0.39$ & $0.34$ & $0.07$ & $0.06$ & $0.00$ & $0.00$ & $0.00$ & $0.00$ & $0.06$ & $0.06$ & $0.07$ & $0.06$ \\
\dset{soybean} & $9.84$ & $1.60$ & $2.78$ & $0.36$ & $2.99$ & $0.43$ & $9.61$ & $1.57$ & $3.07$ & $0.52$ & $3.97$ & $0.55$ & $11.41$ & $1.85$ & $\mathbf{2.13}$ & $\mathbf{0.29}$ & $\mathbf{2.08}$ & $\mathbf{0.30}^*$ & $5.80$ & $0.70$ & $6.66$ & $1.00$ \\
\dset{arrhythmia} & $6.46$ & $1.23$ & $0.55$ & $0.08$ & $0.63$ & $0.08$ & $8.69$ & $1.78$ & $0.57$ & $0.19$ & $0.55$ & $0.17$ & $7.32$ & $1.48$ & $\mathbf{0.36}$ & $\mathbf{0.05}^*$ & $\mathbf{0.37}$ & $\mathbf{0.05}$ & $\mathbf{0.40}$ & $\mathbf{0.06}$ & $\mathbf{0.40}$ & $\mathbf{0.06}$ \\
\dset{optdigits} & $5.33$ & $0.34$ & $5.64$ & $0.26$ & $\mathbf{4.90}$ & $\mathbf{0.35}^*$ & $6.23$ & $0.34$ & $7.67$ & $0.43$ & $6.57$ & $0.35$ & $\mathbf{4.98}$ & $\mathbf{0.26}$ & $6.12$ & $0.32$ & $\mathbf{5.23}$ & $\mathbf{0.31}$ & $\mathbf{4.92}$ & $\mathbf{0.27}$ & $\mathbf{4.92}$ & $\mathbf{0.27}$ \\
\dset{mfeat} & $\mathbf{7.99}$ & $\mathbf{0.55}$ & $9.27$ & $0.74$ & $\mathbf{7.56}$ & $\mathbf{0.55}^*$ & $11.74$ & $0.76$ & $11.23$ & $0.89$ & $10.87$ & $0.83$ & $8.74$ & $0.59$ & $8.36$ & $0.61$ & $8.70$ & $0.64$ & $8.59$ & $0.60$ & $8.54$ & $0.60$ \\
\dset{pendigit} & $1.99$ & $0.11$ & $2.46$ & $0.12$ & $\mathbf{1.88}$ & $\mathbf{0.09}$ & $2.12$ & $0.11$ & $2.36$ & $0.11$ & $2.43$ & $0.19$ & $\mathbf{1.88}$ & $\mathbf{0.10}$ & $1.95$ & $0.08$ & $1.95$ & $0.08$ & $\mathbf{1.85}$ & $\mathbf{0.12}$ & $\mathbf{1.80}$ & $\mathbf{0.11}^*$ \\
  \end{tabular}
  }}\\
  \hspace{-3cm}
 
  (those with the lowest mean are marked with *; those within one standard error of the lowest one are in bold)    
\vskip -0.1in
\end{table*}

\begin{table*}[h]
  \caption{average test error (\%) on benchmark data sets for inconsistant cost}
  \centering
  \label{tbl:softresult2}
  {
  \scalebox{0.53}{
  \begin{tabular}{
      c|
      r@{$\pm$}l@{\hspace{7pt}}
      r@{$\pm$}l@{\hspace{7pt}}
      r@{$\pm$}l@{\hspace{7pt}}
      r@{$\pm$}l@{\hspace{7pt}}
      r@{$\pm$}l@{\hspace{7pt}}
       r@{$\pm$}l@{\hspace{7pt}}
        r@{$\pm$}l@{\hspace{7pt}}
            r@{$\pm$}l@{\hspace{7pt}}
        r@{$\pm$}l@{\hspace{7pt}}
                  r@{$\pm$}l@{\hspace{7pt}}
        r@{$\pm$}l@{\hspace{7pt}}
    }
  data set  & 
   \multicolumn{2}{c}{OVA} &
    \multicolumn{2}{c}{OSR} &
    \multicolumn{2}{c}{soft-OSR} &
      \multicolumn{2}{c}{FT} &
     \multicolumn{2}{c}{CSFT} &     
    \multicolumn{2}{c}{soft-CSFT}&
    \multicolumn{2}{c}{OVO/ZL} &
    \multicolumn{2}{c}{CSOVO} &   
    \multicolumn{2}{c}{soft-CSOVO} &  
    \multicolumn{2}{c}{CSZL} &     
    \multicolumn{2}{c}{soft-CSZL}\\
    \hline
     \dset{iris} & $\mathbf{4.21}$ & $\mathbf{0.78}^*$ & $6.71$ & $0.98$ & $\mathbf{4.74}$ & $\mathbf{0.73}$ & $\mathbf{4.61}$ & $\mathbf{0.79}$ & $7.11$ & $1.24$ & $\mathbf{4.47}$ & $\mathbf{0.81}$ & $\mathbf{4.74}$ & $\mathbf{0.80}$ & $10.66$ & $2.32$ & $5.26$ & $0.72$ & $8.03$ & $1.30$ & $5.39$ & $0.60$ \\
\dset{wine} & $\mathbf{1.78}$ & $\mathbf{0.43}$ & $4.00$ & $0.62$ & $\mathbf{2.00}$ & $\mathbf{0.41}$ & $2.22$ & $0.47$ & $\mathbf{1.67}$ & $\mathbf{0.44}^*$ & $2.22$ & $0.57$ & $2.11$ & $0.51$ & $\mathbf{1.78}$ & $\mathbf{0.51}$ & $\mathbf{1.78}$ & $\mathbf{0.54}$ & $\mathbf{2.09}$ & $\mathbf{0.48}$ & $2.56$ & $0.50$ \\
\dset{glass} & $\mathbf{28.52}$ & $\mathbf{0.82}^*$ & $32.22$ & $1.11$ & $31.94$ & $1.21$ & $29.81$ & $0.96$ & $39.17$ & $2.35$ & $36.02$ & $2.52$ & $\mathbf{28.89}$ & $\mathbf{0.84}$ & $44.26$ & $2.73$ & $45.28$ & $2.52$ & $33.80$ & $1.73$ & $32.50$ & $1.39$ \\
\dset{vehicle} & $\mathbf{20.66}$ & $\mathbf{0.62}$ & $24.15$ & $0.83$ & $22.78$ & $0.73$ & $\mathbf{20.75}$ & $\mathbf{0.64}$ & $29.88$ & $2.92$ & $30.40$ & $3.04$ & $\mathbf{20.31}$ & $\mathbf{0.67}^*$ & $28.73$ & $2.19$ & $25.14$ & $1.57$ & $24.39$ & $1.68$ & $23.00$ & $1.24$ \\
\dset{vowel} & $\mathbf{1.27}$ & $\mathbf{0.17}^*$ & $5.38$ & $0.47$ & $1.88$ & $0.27$ & $1.94$ & $0.24$ & $6.25$ & $1.43$ & $2.74$ & $0.39$ & $\mathbf{1.29}$ & $\mathbf{0.18}$ & $5.93$ & $0.63$ & $\mathbf{1.43}$ & $\mathbf{0.17}$ & $\mathbf{1.31}$ & $\mathbf{0.18}$ & $\mathbf{1.31}$ & $\mathbf{0.18}$ \\
\dset{segment} & $\mathbf{2.60}$ & $\mathbf{0.16}^*$ & $3.69$ & $0.27$ & $\mathbf{2.76}$ & $\mathbf{0.15}$ & $2.78$ & $0.15$ & $4.30$ & $0.62$ & $3.43$ & $0.35$ & $\mathbf{2.60}$ & $\mathbf{0.15}$ & $5.57$ & $0.95$ & $4.11$ & $0.59$ & $\mathbf{2.67}$ & $\mathbf{0.18}$ & $2.78$ & $0.20$ \\
\dset{dna} & $\mathbf{4.20}$ & $\mathbf{0.14}$ & $6.96$ & $0.65$ & $4.87$ & $0.27$ & $4.81$ & $0.24$ & $9.14$ & $1.52$ & $5.32$ & $0.30$ & $\mathbf{4.19}$ & $\mathbf{0.13}^*$ & $7.90$ & $0.80$ & $5.81$ & $0.85$ & $4.74$ & $0.25$ & $4.39$ & $0.15$ \\
\dset{satimage} & $\mathbf{7.19}$ & $\mathbf{0.10}^*$ & $9.52$ & $0.30$ & $9.01$ & $0.34$ & $7.55$ & $0.11$ & $10.58$ & $0.63$ & $9.85$ & $0.75$ & $\mathbf{7.24}$ & $\mathbf{0.09}$ & $12.55$ & $0.66$ & $12.51$ & $0.68$ & $7.87$ & $0.16$ & $7.99$ & $0.30$ \\
\dset{usps} & $\mathbf{2.19}$ & $\mathbf{0.07}^*$ & $3.82$ & $0.13$ & $2.66$ & $0.11$ & $2.79$ & $0.06$ & $6.26$ & $0.86$ & $3.50$ & $0.10$ & $2.28$ & $0.06$ & $5.27$ & $0.70$ & $3.53$ & $0.17$ & $2.33$ & $0.06$ & $2.27$ & $0.06$ \\
\dset{zoo} & $\mathbf{5.19}$ & $\mathbf{0.83}$ & $15.38$ & $1.61$ & $12.50$ & $1.51$ & $\mathbf{4.81}$ & $\mathbf{0.81}^*$ & $12.69$ & $2.54$ & $8.27$ & $2.26$ & $6.15$ & $1.03$ & $10.77$ & $1.71$ & $8.08$ & $1.74$ & $10.77$ & $2.83$ & $14.04$ & $3.01$ \\
\dset{yeast} & $40.38$ & $0.64$ & $73.76$ & $0.55$ & $73.68$ & $0.55$ & $40.20$ & $0.52$ & $77.02$ & $0.92$ & $76.70$ & $0.81$ & $\mathbf{39.27}$ & $\mathbf{0.56}^*$ & $76.58$ & $0.68$ & $76.70$ & $0.67$ & $75.96$ & $0.70$ & $76.31$ & $0.65$ \\
\dset{pageblock} & $3.22$ & $0.09$ & $39.25$ & $4.36$ & $38.54$ & $4.74$ & $\mathbf{3.10}$ & $\mathbf{0.10}$ & $78.25$ & $6.10$ & $81.82$ & $5.81$ & $\mathbf{3.06}$ & $\mathbf{0.08}^*$ & $76.75$ & $6.18$ & $76.75$ & $6.18$ & $80.51$ & $5.88$ & $83.14$ & $5.56$ \\
\dset{anneal} & $\mathbf{1.40}$ & $\mathbf{0.15}^*$ & $8.78$ & $0.94$ & $6.98$ & $1.13$ & $\mathbf{1.47}$ & $\mathbf{0.17}$ & $11.31$ & $1.94$ & $9.47$ & $4.40$ & $\mathbf{1.51}$ & $\mathbf{0.15}$ & $19.02$ & $4.24$ & $10.60$ & $4.53$ & $9.44$ & $4.50$ & $12.07$ & $6.19$ \\
\dset{solar} & $27.27$ & $0.42$ & $34.83$ & $1.16$ & $35.22$ & $1.75$ & $27.27$ & $0.46$ & $46.15$ & $3.12$ & $43.48$ & $2.85$ & $\mathbf{26.61}$ & $\mathbf{0.43}^*$ & $47.49$ & $3.30$ & $47.83$ & $3.12$ & $41.21$ & $3.30$ & $41.54$ & $3.36$ \\
\dset{splice} & $\mathbf{3.86}$ & $\mathbf{0.15}^*$ & $7.68$ & $1.16$ & $5.21$ & $0.56$ & $4.62$ & $0.18$ & $9.59$ & $1.46$ & $6.52$ & $0.74$ & $\mathbf{3.92}$ & $\mathbf{0.12}$ & $13.34$ & $2.69$ & $8.13$ & $2.60$ & $5.49$ & $0.96$ & $5.34$ & $0.98$ \\
\dset{ecoli} & $15.12$ & $0.99$ & $32.68$ & $1.67$ & $33.63$ & $1.61$ & $16.85$ & $1.14$ & $36.73$ & $2.72$ & $40.89$ & $3.85$ & $\mathbf{14.05}$ & $\mathbf{0.75}^*$ & $37.80$ & $3.30$ & $38.45$ & $3.19$ & $38.57$ & $3.00$ & $38.99$ & $2.99$ \\
\dset{nursery} & $0.11$ & $0.02$ & $33.33$ & $0.17$ & $31.02$ & $1.54$ & $0.32$ & $0.08$ & $33.89$ & $0.44$ & $20.04$ & $3.61$ & $\mathbf{0.02}$ & $\mathbf{0.01}^*$ & $37.62$ & $2.17$ & $3.31$ & $2.21$ & $1.69$ & $1.63$ & $\mathbf{0.02}$ & $\mathbf{0.01}$ \\
\dset{soybean} & $\mathbf{6.55}$ & $\mathbf{0.32}^*$ & $24.53$ & $0.82$ & $21.67$ & $1.42$ & $7.13$ & $0.38$ & $35.41$ & $2.48$ & $28.48$ & $3.40$ & $7.46$ & $0.34$ & $39.06$ & $3.51$ & $40.12$ & $3.76$ & $24.47$ & $3.25$ & $20.50$ & $3.56$ \\
\dset{arrhythmia} & $\mathbf{28.41}$ & $\mathbf{0.93}$ & $66.37$ & $2.25$ & $66.42$ & $2.11$ & $30.40$ & $0.62$ & $88.81$ & $2.47$ & $86.15$ & $3.12$ & $\mathbf{27.92}$ & $\mathbf{0.74}^*$ & $85.18$ & $2.49$ & $83.05$ & $3.37$ & $86.68$ & $2.66$ & $84.87$ & $3.34$ \\
\dset{optdigits} & $1.09$ & $0.06$ & $1.85$ & $0.06$ & $1.15$ & $0.07$ & $1.35$ & $0.05$ & $2.14$ & $0.24$ & $1.55$ & $0.05$ & $\mathbf{1.04}$ & $\mathbf{0.05}^*$ & $2.25$ & $0.09$ & $1.36$ & $0.12$ & $1.09$ & $0.05$ & $\mathbf{1.04}$ & $\mathbf{0.05}$ \\
\dset{mfeat} & $\mathbf{1.69}$ & $\mathbf{0.09}^*$ & $3.10$ & $0.18$ & $1.84$ & $0.11$ & $2.45$ & $0.10$ & $3.89$ & $0.37$ & $2.99$ & $0.38$ & $1.86$ & $0.08$ & $4.32$ & $0.53$ & $2.50$ & $0.22$ & $1.85$ & $0.08$ & $1.90$ & $0.09$ \\
\dset{pendigit} & $\mathbf{0.40}$ & $\mathbf{0.02}$ & $0.85$ & $0.04$ & $\mathbf{0.39}$ & $\mathbf{0.02}$ & $0.45$ & $0.02$ & $0.62$ & $0.04$ & $0.52$ & $0.03$ & $\mathbf{0.38}$ & $\mathbf{0.02}^*$ & $0.65$ & $0.03$ & $0.42$ & $0.02$ & $\mathbf{0.40}$ & $\mathbf{0.02}$ & $\mathbf{0.39}$ & $\mathbf{0.02}$ \\
  \end{tabular}
  }}\\
  
  (those with the lowest mean are marked with *; those within one standard error of the lowest one are in bold)    
\vskip -0.1in
\end{table*}

From Table~\ref{tbl:softresult},  soft-OSR and soft-CSOVO usually result in the lowest test cost. Most importantly, soft-OSR is among the best algorithms (bold) on 17 of the 22 data sets, and achieves the lowest cost on 8 of them. The follow-ups, OSR and CSOVO, were the state-of-the-art algorithms in cost-sensitive classification and reach promising performance often. Filter-tree and  CSZL algorithms (CSFT, soft-CSFT, CSZL, soft-CSZL) are generally falling behind, and so are the regular classification algorithms (OVA, OVO, FT). The results justify that soft cost-sensitive classification can lead to similar and sometimes even better performance when compared with state-of-art cost-sensitive classification algorithms. 

The experiments from Table~\ref{tbl:softresult} also indicate cost-sensitive classification algorithms are sometimes overfitting in cost. 
For instance, in data set \dset{vowel}, all state-of-the-art cost-sensitive algorithms are inferior to their regular sibling algorithms in cost.
In data set \dset{dna}, although OSR achieves the similar cost to OVA, the two hard cost-sensitive classification algorithms CSOVO and CSFT are worse to OVO and FT, respectively. For these two data sets, soft cost-sensitive algorithms generally perform better than their hard siblings, and can often achieve lower costs than regular algorithms. The results justify the usefulness of soft cost-sensitive classification. 

When we move to Table~\ref{tbl:softresult2}, regular classification algorithms like OVA and OVO generally achieve the lowest test errors. The hard cost-sensitive classification ones result in the highest test errors; soft ones lie in between.

Soft cost-sensitive classification does not improve CSZL significantly  in terms of either the cost or the error rate. In particular, soft-CSZL ties with CSZL in cost on all 22 data sets, and results in lower error rate in only two of the data sets. One possible reason is that CSZL is implicitly ``soft'' in using the cost information when the cost matrix is inconsistent (i.e. CSZL needs to resort to an approximate solution), and readily leads to low error rate. In particular, CSZL (based on weighted OVO) reaches better error rate than  CSOVO on  $16$  of the $22$ data set; Thus, there is less room to improve CSZL with the proposed methodology. We see that there is no harm in using the soft methodology, though, because the hard CSZL is simply a special case of soft-CSZL with $\alpha=0$.
%
%
%
%
%

%
%

\subsubsection{Consistent Cost Matrix}\label{subsec2}



Next, we consider consistent cost. We use the the same data sets and the normalize procedures. The consistent cost matrices are generated as follows: 

Assume the class number is $K$. We first randomly generate a $K$-dimensional  vector that contains increasing components within $[0, 1]$. We then use those values as solutions of the linear system that  CSZL solves. Then, those components become weights of classes. We associate higher weights to the less frequent classes. The upper triangular of cost   matrix $\CM(k,y), \forall y > k$, can then be uniquely determined from the linear system; we generate the lower triangular of cost  matrix $\CM(y,k), \forall y > k$ from the uniformly  sampled $\left[0, \frac{|\{n:y_n=y\}|}{|\{n:y_n=k\}|} \right]$ and set 
 $\CM(k,k)$  to zero.


Table \ref{tbl:softconresult1} and Table \ref{tbl:softconresult2} are the results when the cost is consistent for CSZL. The results are similar to the results for inconsistent cost. soft-CSOVO is among the best algorithms (bold)  on 18 of the 22 data sets in terms of the cost, followed by soft-OSR, OSR and CSOVO. Filter-tree, CSZL and regular classification algorithms are falling behind. The results again justify that soft cost-sensitive classification could head to better performance when compared with state-of-art cost-sensitive classification algorithms.

From Table \ref{tbl:softconresult1} and Table \ref{tbl:softconresult2}, we observe that soft cost-sensitive classification still could not improve CSZL much in error rate. Note that even when the cost is consistent, the modified cost in (\ref{softeq}) is almost always inconsistent for CSZL when $\alpha > 0$. Such a phase change could be why soft-CSZL does not lead to much improvement, but it is usually no worse than hard CSZL, either. 

Mostly (especially for CSOVO and OSR), soft cost-sensitive classification is better than the regular sibling in terms of the cost, the major criterion. It is similar to (sometimes better than) the hard sibling in terms of the cost, and usually better in terms of the error. We further justify the claims above by comparing the average test cost between soft cost-sensitive classification algorithms with their corresponding siblings using a pairwise one-tailed $t$-test of significance  level $0.1$, as shown in Table~\ref{tbl:pt01_test_regular} for inconsistent cost and Table~\ref{tbl:pt011_test_regular} for consistent cost. The results of these two cost are very similar: for each family of algorithms (OVA, OVO/ZL or FT), soft cost-sensitive classification algorithms  are generally among the best of the three, and are significantly better than their regular siblings (except CSZL).

Table~\ref{tbl:pt01_test} and Table \ref{tbl:pt0111_test} shows the same $t$-test for comparing the test errors between soft cost-sensitive classification algorithms and their hard siblings in inconsistent and consistent costs, respectively.  For inconsistent cost, we see that soft-OSR improves OSR  on $16$ of the $22$ data sets in terms of the test error; soft-CSOVO improves CSOVO on $13$ of the $22$; soft-CSFT improves CSFT on $14$ of the $22$; soft-CSZL improves CSZL on $2$ of the $22$. For consistent cost, we see that soft-OSR improves OSR on $14$ of the $22$ data sets in terms of the test error; soft-CSOVO improves CSOVO on $13$ of the $22$; soft-CSFT improves CSFT on $17$ of the $22$; soft-CSZL improves CSZL on $3$ of the $22$. Given the similar test cost between soft and hard cost-sensitive classification algorithms in Table~\ref{tbl:pt01_test_regular}, the significant improvements on the test error justify that soft cost-sensitive classification algorithms are better choices for practical applications.

\begin{table*}[h]
  \caption{average test cost ($\cdot 10^{-3}$) on benchmark data sets for consistant cost\protect\linebreak (note that the regular sibling of CSZL is also OVO)}
  \centering
  \label{tbl:softconresult1}
  {
  \scalebox{0.51}{
  \begin{tabular}{
      c|
      r@{$\pm$}l@{\hspace{7pt}}
      r@{$\pm$}l@{\hspace{7pt}}
      r@{$\pm$}l@{\hspace{7pt}}
      r@{$\pm$}l@{\hspace{7pt}}
      r@{$\pm$}l@{\hspace{7pt}}
       r@{$\pm$}l@{\hspace{7pt}}
        r@{$\pm$}l@{\hspace{7pt}}
            r@{$\pm$}l@{\hspace{7pt}}
        r@{$\pm$}l@{\hspace{7pt}}
                  r@{$\pm$}l@{\hspace{7pt}}
        r@{$\pm$}l@{\hspace{7pt}}
    }
  data set  & 
   \multicolumn{2}{c}{OVA} &
    \multicolumn{2}{c}{OSR} &
    \multicolumn{2}{c}{soft-OSR} &
      \multicolumn{2}{c}{FT} &
     \multicolumn{2}{c}{CSFT} &     
    \multicolumn{2}{c}{soft-CSFT}&
    \multicolumn{2}{c}{OVO/ZL} &
    \multicolumn{2}{c}{CSOVO} &   
    \multicolumn{2}{c}{soft-CSOVO} &  
    \multicolumn{2}{c}{CSZL} &     
    \multicolumn{2}{c}{soft-CSZL}\\
    \hline
     \dset{iris} & $46.03$ & $5.11$ & $\mathbf{38.05}$ & $\mathbf{6.12}$ & $\mathbf{37.73}$ & $\mathbf{5.36}^*$ & $44.92$ & $4.25$ & $\mathbf{40.03}$ & $\mathbf{4.50}$ & $46.13$ & $4.90$ & $45.17$ & $3.80$ & $\mathbf{40.03}$ & $\mathbf{4.50}$ & $\mathbf{42.21}$ & $\mathbf{5.07}$ & $\mathbf{42.79}$ & $\mathbf{4.04}$ & $\mathbf{41.00}$ & $\mathbf{4.48}$ \\
\dset{wine} & $\mathbf{11.18}$ & $\mathbf{2.36}$ & $16.59$ & $2.12$ & $\mathbf{12.24}$ & $\mathbf{2.91}$ & $\mathbf{12.04}$ & $\mathbf{2.42}$ & $13.99$ & $1.98$ & $\mathbf{11.45}$ & $\mathbf{2.62}$ & $\mathbf{10.88}$ & $\mathbf{2.53}$ & $\mathbf{11.40}$ & $\mathbf{1.94}$ & $\mathbf{10.61}$ & $\mathbf{2.00}$ & $\mathbf{9.63}$ & $\mathbf{2.66}^*$ & $13.43$ & $2.59$ \\
\dset{glass} & $100.14$ & $9.30$ & $\mathbf{86.57}$ & $\mathbf{8.73}^*$ & $\mathbf{91.88}$ & $\mathbf{9.49}$ & $100.27$ & $8.93$ & $\mathbf{93.40}$ & $\mathbf{9.59}$ & $\mathbf{94.38}$ & $\mathbf{9.01}$ & $106.84$ & $10.59$ & $\mathbf{91.37}$ & $\mathbf{10.52}$ & $\mathbf{91.29}$ & $\mathbf{9.74}$ & $99.10$ & $9.92$ & $\mathbf{93.19}$ & $\mathbf{9.71}$ \\
\dset{vehicle} & $105.20$ & $3.98$ & $103.85$ & $4.61$ & $101.81$ & $4.51$ & $112.48$ & $7.71$ & $104.18$ & $6.14$ & $98.77$ & $5.32$ & $99.28$ & $3.82$ & $102.76$ & $6.75$ & $\mathbf{93.30}$ & $\mathbf{5.06}^*$ & $99.37$ & $3.85$ & $99.26$ & $3.92$ \\
\dset{vowel} & $6.99$ & $1.29$ & $10.98$ & $1.12$ & $6.73$ & $1.26$ & $9.81$ & $1.34$ & $12.61$ & $1.75$ & $10.32$ & $1.40$ & $\mathbf{5.49}$ & $\mathbf{1.01}$ & $10.16$ & $1.28$ & $\mathbf{5.74}$ & $\mathbf{1.04}$ & $\mathbf{5.21}$ & $\mathbf{1.01}$ & $\mathbf{5.04}$ & $\mathbf{1.00}^*$ \\
\dset{segment} & $\mathbf{12.18}$ & $\mathbf{1.17}$ & $12.72$ & $1.21$ & $\mathbf{12.31}$ & $\mathbf{1.18}$ & $\mathbf{12.48}$ & $\mathbf{1.21}$ & $14.29$ & $1.33$ & $\mathbf{12.55}$ & $\mathbf{1.28}$ & $\mathbf{11.58}$ & $\mathbf{1.04}^*$ & $\mathbf{12.42}$ & $\mathbf{1.22}$ & $\mathbf{11.60}$ & $\mathbf{1.07}$ & $\mathbf{11.63}$ & $\mathbf{1.04}$ & $\mathbf{11.63}$ & $\mathbf{1.04}$ \\
\dset{dna} & $\mathbf{19.84}$ & $\mathbf{1.31}$ & $23.05$ & $1.43$ & $\mathbf{19.80}$ & $\mathbf{1.37}^*$ & $25.27$ & $1.61$ & $25.78$ & $2.05$ & $23.91$ & $1.61$ & $\mathbf{20.69}$ & $\mathbf{1.42}$ & $25.86$ & $2.02$ & $\mathbf{20.97}$ & $\mathbf{1.46}$ & $\mathbf{20.45}$ & $\mathbf{1.37}$ & $\mathbf{20.93}$ & $\mathbf{1.49}$ \\
\dset{satimage} & $30.32$ & $2.10$ & $\mathbf{29.23}$ & $\mathbf{2.11}$ & $\mathbf{29.08}$ & $\mathbf{2.10}$ & $31.40$ & $2.07$ & $30.02$ & $2.07$ & $30.53$ & $2.26$ & $30.92$ & $2.07$ & $\mathbf{28.02}$ & $\mathbf{2.02}$ & $\mathbf{27.97}$ & $\mathbf{2.05}^*$ & $31.03$ & $2.04$ & $30.99$ & $2.04$ \\
\dset{usps} & $6.15$ & $0.25$ & $6.05$ & $0.21$ & $\mathbf{5.71}$ & $\mathbf{0.22}^*$ & $9.05$ & $0.29$ & $7.41$ & $0.31$ & $7.07$ & $0.35$ & $6.38$ & $0.28$ & $6.01$ & $0.31$ & $\mathbf{5.91}$ & $\mathbf{0.29}$ & $6.35$ & $0.29$ & $6.32$ & $0.28$ \\
\dset{zoo} & $12.95$ & $2.81$ & $11.57$ & $1.71$ & $10.73$ & $2.41$ & $\mathbf{7.29}$ & $\mathbf{1.80}^*$ & $9.91$ & $2.26$ & $11.25$ & $2.74$ & $11.25$ & $2.45$ & $\mathbf{7.97}$ & $\mathbf{1.72}$ & $9.23$ & $2.38$ & $12.46$ & $2.60$ & $11.79$ & $2.68$ \\
\dset{yeast} & $8.42$ & $0.69$ & $7.34$ & $0.82$ & $7.40$ & $0.80$ & $27.29$ & $2.62$ & $7.79$ & $0.85$ & $7.53$ & $0.75$ & $8.06$ & $0.67$ & $\mathbf{6.51}$ & $\mathbf{0.76}^*$ & $\mathbf{6.66}$ & $\mathbf{0.75}$ & $8.20$ & $0.73$ & $8.20$ & $0.74$ \\
\dset{pageblock} & $7.76$ & $0.81$ & $5.85$ & $0.82$ & $6.02$ & $0.79$ & $\mathbf{4.70}$ & $\mathbf{0.49}^*$ & $7.32$ & $0.95$ & $6.59$ & $0.83$ & $7.52$ & $0.81$ & $5.83$ & $0.80$ & $5.83$ & $0.83$ & $7.81$ & $0.80$ & $7.57$ & $0.82$ \\
\dset{anneal} & $4.73$ & $0.91$ & $4.13$ & $0.57$ & $\mathbf{3.67}$ & $\mathbf{0.53}$ & $5.77$ & $1.04$ & $\mathbf{3.48}$ & $\mathbf{0.57}$ & $\mathbf{3.57}$ & $\mathbf{0.52}$ & $3.92$ & $0.65$ & $\mathbf{3.30}$ & $\mathbf{0.49}$ & $\mathbf{3.29}$ & $\mathbf{0.51}^*$ & $3.92$ & $0.65$ & $3.93$ & $0.65$ \\
\dset{solar} & $54.51$ & $2.51$ & $43.53$ & $2.69$ & $43.66$ & $2.84$ & $49.67$ & $4.40$ & $46.85$ & $3.65$ & $44.07$ & $3.32$ & $58.93$ & $2.90$ & $\mathbf{35.82}$ & $\mathbf{2.69}$ & $\mathbf{35.68}$ & $\mathbf{2.70}^*$ & $56.12$ & $3.00$ & $55.28$ & $2.64$ \\
\dset{splice} & $\mathbf{19.67}$ & $\mathbf{1.06}$ & $25.08$ & $1.35$ & $\mathbf{19.62}$ & $\mathbf{0.98}$ & $\mathbf{18.86}$ & $\mathbf{1.73}^*$ & $28.83$ & $1.82$ & $22.64$ & $1.37$ & $\mathbf{19.87}$ & $\mathbf{1.04}$ & $29.36$ & $2.01$ & $\mathbf{19.73}$ & $\mathbf{0.96}$ & $\mathbf{20.05}$ & $\mathbf{1.03}$ & $\mathbf{19.91}$ & $\mathbf{1.07}$ \\
\dset{ecoli} & $4.84$ & $0.54$ & $\mathbf{4.67}$ & $\mathbf{0.60}$ & $\mathbf{4.20}$ & $\mathbf{0.53}^*$ & $\mathbf{4.28}$ & $\mathbf{0.80}$ & $5.89$ & $0.60$ & $5.76$ & $0.86$ & $\mathbf{4.46}$ & $\mathbf{0.52}$ & $5.11$ & $0.49$ & $\mathbf{4.51}$ & $\mathbf{0.43}$ & $\mathbf{4.44}$ & $\mathbf{0.51}$ & $\mathbf{4.45}$ & $\mathbf{0.51}$ \\
\dset{nursery} & $0.01$ & $0.00$ & $0.04$ & $0.01$ & $\mathbf{0.01}$ & $\mathbf{0.00}$ & $1.42$ & $0.45$ & $0.01$ & $0.00$ & $0.02$ & $0.01$ & $\mathbf{0.01}$ & $\mathbf{0.00}^*$ & $0.01$ & $0.00$ & $\mathbf{0.01}$ & $\mathbf{0.00}$ & $\mathbf{0.01}$ & $\mathbf{0.00}$ & $\mathbf{0.01}$ & $\mathbf{0.00}$ \\
\dset{soybean} & $4.17$ & $0.48$ & $2.83$ & $0.29$ & $3.39$ & $0.40$ & $\mathbf{1.63}$ & $\mathbf{0.07}^*$ & $3.95$ & $0.50$ & $4.59$ & $0.48$ & $5.02$ & $0.66$ & $2.23$ & $0.24$ & $2.23$ & $0.26$ & $4.53$ & $0.61$ & $4.48$ & $0.58$ \\
\dset{arrhythmia} & $24.35$ & $2.63$ & $\mathbf{8.72}$ & $\mathbf{1.71}$ & $\mathbf{8.68}$ & $\mathbf{1.68}$ & $\mathbf{8.69}$ & $\mathbf{1.78}$ & $14.58$ & $2.11$ & $14.39$ & $2.58$ & $26.21$ & $2.90$ & $\mathbf{8.13}$ & $\mathbf{1.77}$ & $\mathbf{7.96}$ & $\mathbf{1.62}^*$ & $25.27$ & $3.06$ & $25.18$ & $3.10$ \\
\dset{optdigits} & $4.24$ & $0.24$ & $4.24$ & $0.24$ & $\mathbf{3.94}$ & $\mathbf{0.26}^*$ & $6.23$ & $0.34$ & $5.85$ & $0.38$ & $5.59$ & $0.36$ & $\mathbf{4.08}$ & $\mathbf{0.27}$ & $4.32$ & $0.21$ & $\mathbf{4.14}$ & $\mathbf{0.24}$ & $\mathbf{4.18}$ & $\mathbf{0.34}$ & $\mathbf{4.10}$ & $\mathbf{0.28}$ \\
\dset{mfeat} & $\mathbf{7.32}$ & $\mathbf{0.50}$ & $7.53$ & $0.58$ & $\mathbf{6.90}$ & $\mathbf{0.56}^*$ & $11.74$ & $0.76$ & $9.49$ & $0.61$ & $9.15$ & $0.57$ & $8.03$ & $0.56$ & $7.55$ & $0.54$ & $7.78$ & $0.58$ & $8.00$ & $0.55$ & $8.21$ & $0.59$ \\
\dset{pendigit} & $1.98$ & $0.13$ & $2.22$ & $0.09$ & $\mathbf{1.85}$ & $\mathbf{0.12}$ & $2.24$ & $0.13$ & $2.38$ & $0.15$ & $2.27$ & $0.11$ & $\mathbf{1.85}$ & $\mathbf{0.10}$ & $2.09$ & $0.10$ & $\mathbf{1.87}$ & $\mathbf{0.10}$ & $\mathbf{1.83}$ & $\mathbf{0.12}$ & $\mathbf{1.78}$ & $\mathbf{0.11}^*$ \\
  \end{tabular}
  }}\\
  
  (those with the lowest mean are marked with *; those within one standard error of the lowest one are in bold)    
\vskip -0.1in
\end{table*}

\begin{table*}[h]
  \caption{average test error (\%) on benchmark data sets for consistent cost}
  \centering
  \label{tbl:softconresult2}
  {
  \scalebox{0.53}{
  \begin{tabular}{
      c|
      r@{$\pm$}l@{\hspace{7pt}}
      r@{$\pm$}l@{\hspace{7pt}}
      r@{$\pm$}l@{\hspace{7pt}}
      r@{$\pm$}l@{\hspace{7pt}}
      r@{$\pm$}l@{\hspace{7pt}}
       r@{$\pm$}l@{\hspace{7pt}}
        r@{$\pm$}l@{\hspace{7pt}}
            r@{$\pm$}l@{\hspace{7pt}}
        r@{$\pm$}l@{\hspace{7pt}}
                  r@{$\pm$}l@{\hspace{7pt}}
        r@{$\pm$}l@{\hspace{7pt}}
    }
  data set  & 
   \multicolumn{2}{c}{OVA} &
    \multicolumn{2}{c}{OSR} &
    \multicolumn{2}{c}{soft-OSR} &
      \multicolumn{2}{c}{FT} &
     \multicolumn{2}{c}{CSFT} &     
    \multicolumn{2}{c}{soft-CSFT}&
    \multicolumn{2}{c}{OVO/ZL} &
    \multicolumn{2}{c}{CSOVO} &   
    \multicolumn{2}{c}{soft-CSOVO} &  
    \multicolumn{2}{c}{CSZL} &     
    \multicolumn{2}{c}{soft-CSZL}\\
    \hline
     \dset{iris} & $5.66$ & $0.60$ & $\mathbf{4.74}$ & $\mathbf{0.69}^*$ & $\mathbf{4.74}$ & $\mathbf{0.63}$ & $5.53$ & $0.52$ & $\mathbf{5.00}$ & $\mathbf{0.56}$ & $5.66$ & $0.57$ & $5.53$ & $0.49$ & $\mathbf{5.00}$ & $\mathbf{0.56}$ & $\mathbf{5.26}$ & $\mathbf{0.62}$ & $\mathbf{5.39}$ & $\mathbf{0.57}$ & $\mathbf{5.13}$ & $\mathbf{0.57}$ \\
\dset{wine} & $\mathbf{1.78}$ & $\mathbf{0.37}$ & $3.33$ & $0.43$ & $2.11$ & $0.46$ & $2.11$ & $0.46$ & $2.33$ & $0.33$ & $2.00$ & $0.41$ & $\mathbf{1.78}$ & $\mathbf{0.40}$ & $2.11$ & $0.33$ & $\mathbf{1.89}$ & $\mathbf{0.32}$ & $\mathbf{1.56}$ & $\mathbf{0.39}^*$ & $2.00$ & $0.38$ \\
\dset{glass} & $\mathbf{30.65}$ & $\mathbf{1.31}$ & $34.26$ & $1.84$ & $32.78$ & $1.53$ & $\mathbf{30.74}$ & $\mathbf{1.22}$ & $42.87$ & $3.56$ & $36.94$ & $2.83$ & $34.54$ & $3.52$ & $40.56$ & $3.45$ & $39.91$ & $3.55$ & $\mathbf{30.83}$ & $\mathbf{1.47}$ & $\mathbf{30.28}$ & $\mathbf{1.48}^*$ \\
\dset{vehicle} & $19.62$ & $0.54$ & $22.97$ & $1.42$ & $22.48$ & $1.36$ & $20.75$ & $0.64$ & $24.79$ & $2.17$ & $23.09$ & $2.29$ & $\mathbf{18.56}$ & $\mathbf{0.60}$ & $32.10$ & $2.85$ & $24.98$ & $2.72$ & $\mathbf{18.51}$ & $\mathbf{0.61}^*$ & $19.62$ & $0.68$ \\
\dset{vowel} & $1.75$ & $0.27$ & $5.56$ & $0.67$ & $1.83$ & $0.28$ & $2.38$ & $0.21$ & $7.46$ & $1.38$ & $3.10$ & $0.31$ & $\mathbf{1.45}$ & $\mathbf{0.19}$ & $9.27$ & $1.52$ & $1.75$ & $0.30$ & $\mathbf{1.39}$ & $\mathbf{0.18}$ & $\mathbf{1.35}$ & $\mathbf{0.18}^*$ \\
\dset{segment} & $\mathbf{2.34}$ & $\mathbf{0.11}$ & $3.04$ & $0.13$ & $2.51$ & $0.12$ & $2.44$ & $0.13$ & $3.49$ & $0.28$ & $3.04$ & $0.37$ & $\mathbf{2.26}$ & $\mathbf{0.10}$ & $3.83$ & $0.47$ & $2.98$ & $0.35$ & $\mathbf{2.25}$ & $\mathbf{0.10}^*$ & $\mathbf{2.28}$ & $\mathbf{0.09}$ \\
\dset{dna} & $\mathbf{4.10}$ & $\mathbf{0.13}^*$ & $9.22$ & $1.86$ & $4.47$ & $0.25$ & $5.12$ & $0.29$ & $12.50$ & $3.41$ & $6.34$ & $1.06$ & $4.34$ & $0.14$ & $10.35$ & $2.33$ & $8.34$ & $2.59$ & $4.31$ & $0.16$ & $4.46$ & $0.20$ \\
\dset{satimage} & $\mathbf{7.34}$ & $\mathbf{0.09}^*$ & $8.89$ & $0.23$ & $8.33$ & $0.30$ & $7.64$ & $0.09$ & $10.78$ & $0.70$ & $9.52$ & $0.61$ & $7.51$ & $0.08$ & $12.60$ & $0.84$ & $12.38$ & $0.90$ & $7.52$ & $0.08$ & $7.55$ & $0.09$ \\
\dset{usps} & $\mathbf{2.25}$ & $\mathbf{0.06}^*$ & $4.18$ & $0.29$ & $2.53$ & $0.13$ & $2.79$ & $0.06$ & $6.00$ & $0.91$ & $3.33$ & $0.10$ & $2.35$ & $0.06$ & $6.51$ & $1.28$ & $5.35$ & $1.33$ & $2.33$ & $0.06$ & $2.32$ & $0.06$ \\
\dset{zoo} & $\mathbf{5.19}$ & $\mathbf{0.83}$ & $7.69$ & $1.12$ & $\mathbf{5.19}$ & $\mathbf{1.13}$ & $6.15$ & $1.03$ & $5.58$ & $1.35$ & $\mathbf{4.62}$ & $\mathbf{0.89}^*$ & $\mathbf{5.00}$ & $\mathbf{0.72}$ & $6.54$ & $0.98$ & $\mathbf{5.19}$ & $\mathbf{0.78}$ & $5.58$ & $0.92$ & $\mathbf{5.00}$ & $\mathbf{0.77}$ \\
\dset{yeast} & $40.43$ & $0.56$ & $46.59$ & $1.12$ & $45.82$ & $1.16$ & $\mathbf{39.91}$ & $\mathbf{0.58}$ & $56.33$ & $2.09$ & $51.64$ & $2.31$ & $\mathbf{39.47}$ & $\mathbf{0.51}^*$ & $53.38$ & $1.39$ & $53.45$ & $1.40$ & $41.00$ & $0.46$ & $41.00$ & $0.46$ \\
\dset{pageblock} & $3.19$ & $0.09$ & $4.49$ & $0.25$ & $4.67$ & $0.30$ & $3.22$ & $0.10$ & $6.47$ & $0.60$ & $6.45$ & $0.88$ & $\mathbf{3.06}$ & $\mathbf{0.10}^*$ & $7.29$ & $0.60$ & $7.43$ & $0.67$ & $\mathbf{3.14}$ & $\mathbf{0.09}$ & $\mathbf{3.13}$ & $\mathbf{0.09}$ \\
\dset{anneal} & $3.17$ & $1.75$ & $5.56$ & $0.77$ & $3.27$ & $0.71$ & $\mathbf{1.49}$ & $\mathbf{0.15}$ & $4.11$ & $1.49$ & $3.44$ & $1.13$ & $\mathbf{1.49}$ & $\mathbf{0.15}$ & $4.31$ & $1.04$ & $2.47$ & $0.74$ & $\mathbf{1.47}$ & $\mathbf{0.16}^*$ & $\mathbf{1.51}$ & $\mathbf{0.17}$ \\
\dset{solar} & $27.28$ & $0.46$ & $31.25$ & $0.92$ & $30.23$ & $1.03$ & $\mathbf{26.44}$ & $\mathbf{0.38}^*$ & $39.32$ & $1.74$ & $41.41$ & $2.37$ & $26.90$ & $0.45$ & $41.85$ & $1.84$ & $42.34$ & $1.91$ & $27.44$ & $0.53$ & $28.23$ & $0.46$ \\
\dset{splice} & $\mathbf{3.87}$ & $\mathbf{0.15}^*$ & $6.60$ & $0.59$ & $4.08$ & $0.19$ & $4.47$ & $0.24$ & $6.99$ & $0.63$ & $4.81$ & $0.32$ & $\mathbf{3.87}$ & $\mathbf{0.12}$ & $6.74$ & $0.53$ & $\mathbf{3.92}$ & $\mathbf{0.13}$ & $\mathbf{3.88}$ & $\mathbf{0.13}$ & $\mathbf{3.91}$ & $\mathbf{0.13}$ \\
\dset{ecoli} & $15.12$ & $0.99$ & $21.61$ & $2.79$ & $18.81$ & $2.80$ & $16.85$ & $0.72$ & $28.57$ & $3.90$ & $23.10$ & $2.90$ & $\mathbf{13.99}$ & $\mathbf{0.76}$ & $32.86$ & $4.95$ & $25.00$ & $4.52$ & $\mathbf{13.93}$ & $\mathbf{0.73}^*$ & $\mathbf{13.99}$ & $\mathbf{0.76}$ \\
\dset{nursery} & $0.09$ & $0.02$ & $3.44$ & $0.34$ & $0.04$ & $0.01$ & $0.32$ & $0.08$ & $6.23$ & $2.01$ & $0.35$ & $0.07$ & $\mathbf{0.01}$ & $\mathbf{0.00}^*$ & $6.55$ & $2.30$ & $\mathbf{0.02}$ & $\mathbf{0.00}$ & $0.06$ & $0.05$ & $\mathbf{0.01}$ & $\mathbf{0.00}$ \\
\dset{soybean} & $\mathbf{6.43}$ & $\mathbf{0.30}^*$ & $11.61$ & $1.52$ & $10.18$ & $1.62$ & $7.02$ & $0.28$ & $22.89$ & $2.50$ & $14.42$ & $2.76$ & $7.31$ & $0.36$ & $25.99$ & $3.19$ & $23.95$ & $3.22$ & $8.74$ & $1.12$ & $8.65$ & $1.14$ \\
\dset{arrhythmia} & $\mathbf{28.67}$ & $\mathbf{0.89}$ & $69.11$ & $2.59$ & $68.41$ & $3.11$ & $30.40$ & $0.62$ & $73.14$ & $5.96$ & $66.28$ & $6.31$ & $\mathbf{28.05}$ & $\mathbf{0.74}^*$ & $86.59$ & $3.62$ & $84.38$ & $4.41$ & $32.70$ & $3.50$ & $32.88$ & $3.48$ \\
\dset{optdigits} & $\mathbf{1.09}$ & $\mathbf{0.06}$ & $2.48$ & $0.14$ & $1.22$ & $0.06$ & $1.35$ & $0.05$ & $3.35$ & $1.08$ & $1.74$ & $0.08$ & $\mathbf{1.05}$ & $\mathbf{0.05}^*$ & $3.72$ & $0.59$ & $1.58$ & $0.15$ & $\mathbf{1.05}$ & $\mathbf{0.06}$ & $\mathbf{1.05}$ & $\mathbf{0.05}$ \\
\dset{mfeat} & $\mathbf{1.69}$ & $\mathbf{0.09}^*$ & $3.45$ & $0.22$ & $1.93$ & $0.10$ & $2.45$ & $0.10$ & $4.85$ & $1.00$ & $2.82$ & $0.13$ & $1.81$ & $0.08$ & $4.62$ & $0.83$ & $2.49$ & $0.18$ & $1.85$ & $0.08$ & $1.86$ & $0.08$ \\
\dset{pendigit} & $0.40$ & $0.02$ & $0.89$ & $0.04$ & $0.41$ & $0.02$ & $0.47$ & $0.03$ & $0.73$ & $0.05$ & $0.54$ & $0.03$ & $\mathbf{0.38}$ & $\mathbf{0.02}$ & $0.74$ & $0.05$ & $0.44$ & $0.02$ & $0.43$ & $0.02$ & $\mathbf{0.38}$ & $\mathbf{0.02}^*$ \\
  \end{tabular}
  }}\\
  
  (those with the lowest mean are marked with *; those within one standard error of the lowest one are in bold)    
\vskip -0.1in
\end{table*}

\begin{table}[h]
  \caption{comparisons on the test cost between the algorithms and their soft cost-sensitive classification sibling
    using a pairwise one-tailed $t$-test of significance level $0.05$ in inconsistent cost}
  \vskip 0.1in
  \label{tbl:pt01_test_regular}
  {
    \begin{center}
   \scalebox{0.8}{
  \begin{tabular}{
      c||cc|cc|cc|cc
    }
    data set &  
	{OVA} &
	{OSR} &
	{OVO/ZL} &
	{CSOVO} &
	{FT} &        
	{CSFT} &
	{OVO/ZL} &
	{CSZL} \\
    \hline
    \dset{iris} &$\thickapprox$ & $\thickapprox$ & $\thickapprox$ & $\thickapprox$ & $\thickapprox$ & $\thickapprox$ & $\thickapprox$ & $\thickapprox$  \\  
\dset{wine} &$\thickapprox$ & $\thickapprox$ & $\thickapprox$ & $\thickapprox$ & $\thickapprox$ & $\thickapprox$ & $\thickapprox$ & $\thickapprox$  \\  
\dset{glass} &$\bigcirc$ & $\thickapprox$ & $\thickapprox$ & $\thickapprox$ & $\thickapprox$ & $\thickapprox$ & $\thickapprox$ & $\thickapprox$  \\  
\dset{vehicle} &$\thickapprox$ & $\thickapprox$ & $\thickapprox$ & $\thickapprox$ & $\thickapprox$ & $\thickapprox$ & $\thickapprox$ & $\thickapprox$  \\  
\dset{vowel} &$\thickapprox$ & $\bigcirc$ & $\thickapprox$ & $\thickapprox$ & $\thickapprox$ & $\thickapprox$ & $\thickapprox$ & $\thickapprox$  \\  
\dset{segment} &$\thickapprox$ & $\thickapprox$ & $\thickapprox$ & $\thickapprox$ & $\thickapprox$ & $\thickapprox$ & $\thickapprox$ & $\thickapprox$  \\  
\dset{dna} &$\thickapprox$ & $\thickapprox$ & $\thickapprox$ & $\thickapprox$ & $\thickapprox$ & $\thickapprox$ & $\thickapprox$ & $\thickapprox$  \\  
\dset{satimage} &$\thickapprox$ & $\thickapprox$ & $\thickapprox$ & $\thickapprox$ & $\thickapprox$ & $\thickapprox$ & $\thickapprox$ & $\thickapprox$  \\  
\dset{usps} &$\thickapprox$ & $\bigcirc$ & $\thickapprox$ & $\thickapprox$ & $\thickapprox$ & $\thickapprox$ & $\thickapprox$ & $\thickapprox$  \\  
\dset{zoo} &$\thickapprox$ & $\thickapprox$ & $\bigcirc$ & $\thickapprox$ & $\thickapprox$ & $\thickapprox$ & $\thickapprox$ & $\thickapprox$  \\  
\dset{yeast} &$\bigcirc$ & $\thickapprox$ & $\bigcirc$ & $\thickapprox$ & $\bigcirc$ & $\thickapprox$ & $\bigcirc$ & $\thickapprox$  \\  
\dset{pageblock} &$\bigcirc$ & $\thickapprox$ & $\bigcirc$ & $\thickapprox$ & $\bigcirc$ & $\thickapprox$ & $\bigcirc$ & $\thickapprox$  \\  
\dset{anneal} &$\thickapprox$ & $\thickapprox$ & $\thickapprox$ & $\thickapprox$ & $\thickapprox$ & $\thickapprox$ & $\thickapprox$ & $\thickapprox$  \\  
\dset{solar} &$\bigcirc$ & $\thickapprox$ & $\bigcirc$ & $\thickapprox$ & $\bigcirc$ & $\thickapprox$ & $\bigcirc$ & $\thickapprox$  \\  
\dset{splice} &$\thickapprox$ & $\thickapprox$ & $\thickapprox$ & $\bigcirc$ & $\thickapprox$ & $\thickapprox$ & $\thickapprox$ & $\thickapprox$  \\  
\dset{ecoli} &$\bigcirc$ & $\thickapprox$ & $\bigcirc$ & $\thickapprox$ & $\bigcirc$ & $\thickapprox$ & $\bigcirc$ & $\thickapprox$  \\  
\dset{nursery} &$\bigcirc$ & $\thickapprox$ & $\thickapprox$ & $\thickapprox$ & $\thickapprox$ & $\thickapprox$ & $\thickapprox$ & $\thickapprox$  \\  
\dset{soybean} &$\bigcirc$ & $\thickapprox$ & $\bigcirc$ & $\thickapprox$ & $\bigcirc$ & $\thickapprox$ & $\bigcirc$ & $\thickapprox$  \\  
\dset{arrhythmia} &$\bigcirc$ & $\thickapprox$ & $\bigcirc$ & $\thickapprox$ & $\bigcirc$ & $\thickapprox$ & $\bigcirc$ & $\thickapprox$  \\  
\dset{optdigits} &$\thickapprox$ & $\thickapprox$ & $\thickapprox$ & $\thickapprox$ & $\thickapprox$ & $\thickapprox$ & $\thickapprox$ & $\thickapprox$  \\  
\dset{mfeat} &$\thickapprox$ & $\thickapprox$ & $\thickapprox$ & $\thickapprox$ & $\thickapprox$ & $\thickapprox$ & $\thickapprox$ & $\thickapprox$  \\  
\dset{pendigit} &$\thickapprox$ & $\bigcirc$ & $\thickapprox$ & $\thickapprox$ & $\thickapprox$ & $\thickapprox$ & $\thickapprox$ & $\thickapprox$  \\  
  \end{tabular}
  }
  \begin{tabular}{c@{$\colon$}l}
  $\bigcirc$ & soft cost-sensitive algorithms significantly better\\ 
  $\times$ & soft cost-sensitive  algorithms significantly worse\\ 
  $\thickapprox$ & otherwise
  \end{tabular}

    \end{center}
}
\end{table}

\begin{table}[h]
  \caption{comparison on the test errors between the hard cost-sensitive classification algorithms and their soft sibling
    using a pairwise one-tailed $t$-test of significance level $0.05$ in inconsistent cost}
  \vskip 0.1in
  \label{tbl:pt01_test}
  {
    \begin{center}
   \scalebox{0.85}{
  \begin{tabular}{
      c||c|c|c|c
    }
    data set &  
	{OSR} &
	{CSOVO} &
	{CSFT}&
	{CSZL} \\
    \hline
    \dset{iris} &$\thickapprox$ & $\bigcirc$ & $\thickapprox$ & $\thickapprox$  \\  
\dset{wine} &$\bigcirc$ & $\thickapprox$ & $\thickapprox$ & $\thickapprox$  \\  
\dset{glass} &$\thickapprox$ & $\thickapprox$ & $\thickapprox$ & $\thickapprox$  \\  
\dset{vehicle} &$\thickapprox$ & $\thickapprox$ & $\thickapprox$ & $\thickapprox$  \\  
\dset{vowel} &$\bigcirc$ & $\bigcirc$ & $\bigcirc$ & $\thickapprox$  \\  
\dset{segment} &$\bigcirc$ & $\thickapprox$ & $\thickapprox$ & $\thickapprox$  \\  
\dset{dna} &$\bigcirc$ & $\thickapprox$ & $\bigcirc$ & $\thickapprox$  \\  
\dset{satimage} &$\thickapprox$ & $\thickapprox$ & $\thickapprox$ & $\thickapprox$  \\  
\dset{usps} &$\bigcirc$ & $\bigcirc$ & $\bigcirc$ & $\thickapprox$  \\  
\dset{zoo} &$\thickapprox$ & $\thickapprox$ & $\thickapprox$ & $\thickapprox$  \\  
\dset{yeast} &$\thickapprox$ & $\thickapprox$ & $\thickapprox$ & $\thickapprox$  \\  
\dset{pageblock} &$\thickapprox$ & $\thickapprox$ & $\thickapprox$ & $\thickapprox$  \\  
\dset{anneal} &$\thickapprox$ & $\thickapprox$ & $\thickapprox$ & $\thickapprox$  \\  
\dset{solar} &$\thickapprox$ & $\thickapprox$ & $\thickapprox$ & $\thickapprox$  \\  
\dset{splice} &$\thickapprox$ & $\thickapprox$ & $\thickapprox$ & $\thickapprox$  \\  
\dset{ecoli} &$\thickapprox$ & $\thickapprox$ & $\thickapprox$ & $\thickapprox$  \\  
\dset{nursery} &$\thickapprox$ & $\bigcirc$ & $\bigcirc$ & $\thickapprox$  \\  
\dset{soybean} &$\thickapprox$ & $\thickapprox$ & $\thickapprox$ & $\thickapprox$  \\  
\dset{arrhythmia} &$\thickapprox$ & $\thickapprox$ & $\thickapprox$ & $\thickapprox$  \\  
\dset{optdigits} &$\bigcirc$ & $\bigcirc$ & $\bigcirc$ & $\thickapprox$  \\  
\dset{mfeat} &$\bigcirc$ & $\bigcirc$ & $\thickapprox$ & $\thickapprox$  \\  
\dset{pendigit} &$\bigcirc$ & $\bigcirc$ & $\thickapprox$ & $\thickapprox$  \\
  \end{tabular}
  }
  \begin{tabular}{c@{$\colon$}l}
  $\bigcirc$ & soft cost-sensitive algorithms significantly better\\ 
  $\times$ & soft cost-sensitive  algorithms significantly worse\\ 
  $\thickapprox$ & otherwise
  \end{tabular}
    \end{center}
}
\end{table}

\begin{table}[h]
  \caption{comparisons on the test cost between the algorithms and their soft cost-sensitive classification sibling
    using a pairwise one-tailed $t$-test of significance level $0.05$ in consistent cost}
  \vskip 0.1in
  \label{tbl:pt011_test_regular}
  {
    \begin{center}
   \scalebox{0.8}{
  \begin{tabular}{
      c||cc|cc|cc|cc
    }
    data set &  
	{OVA} &
	{OSR} &
	{OVO/ZL} &
	{CSOVO} &
	{FT} &        
	{CSFT} &
	{OVO/ZL} &
	{CSZL} \\
    \hline
     \dset{iris} &$\thickapprox$ & $\thickapprox$ & $\thickapprox$ & $\thickapprox$ & $\thickapprox$ & $\thickapprox$ & $\thickapprox$ & $\thickapprox$  \\  
\dset{wine} &$\thickapprox$ & $\thickapprox$ & $\thickapprox$ & $\thickapprox$ & $\thickapprox$ & $\thickapprox$ & $\thickapprox$ & $\thickapprox$  \\  
\dset{glass} &$\thickapprox$ & $\thickapprox$ & $\thickapprox$ & $\thickapprox$ & $\thickapprox$ & $\thickapprox$ & $\thickapprox$ & $\thickapprox$  \\  
\dset{vehicle} &$\thickapprox$ & $\thickapprox$ & $\thickapprox$ & $\thickapprox$ & $\thickapprox$ & $\thickapprox$ & $\thickapprox$ & $\thickapprox$  \\  
\dset{vowel} &$\thickapprox$ & $\bigcirc$ & $\thickapprox$ & $\bigcirc$ & $\thickapprox$ & $\thickapprox$ & $\thickapprox$ & $\thickapprox$  \\  
\dset{segment} &$\thickapprox$ & $\thickapprox$ & $\thickapprox$ & $\thickapprox$ & $\thickapprox$ & $\thickapprox$ & $\thickapprox$ & $\thickapprox$  \\  
\dset{dna} &$\thickapprox$ & $\thickapprox$ & $\thickapprox$ & $\thickapprox$ & $\thickapprox$ & $\thickapprox$ & $\thickapprox$ & $\thickapprox$  \\  
\dset{satimage} &$\thickapprox$ & $\thickapprox$ & $\thickapprox$ & $\thickapprox$ & $\thickapprox$ & $\thickapprox$ & $\thickapprox$ & $\thickapprox$  \\  
\dset{usps} &$\thickapprox$ & $\thickapprox$ & $\thickapprox$ & $\thickapprox$ & $\bigcirc$ & $\thickapprox$ & $\thickapprox$ & $\thickapprox$  \\  
\dset{zoo} &$\thickapprox$ & $\thickapprox$ & $\thickapprox$ & $\thickapprox$ & $\thickapprox$ & $\thickapprox$ & $\thickapprox$ & $\thickapprox$  \\  
\dset{yeast} &$\thickapprox$ & $\thickapprox$ & $\thickapprox$ & $\thickapprox$ & $\bigcirc$ & $\thickapprox$ & $\thickapprox$ & $\thickapprox$  \\  
\dset{pageblock} &$\thickapprox$ & $\thickapprox$ & $\thickapprox$ & $\thickapprox$ & $\thickapprox$ & $\thickapprox$ & $\thickapprox$ & $\thickapprox$  \\  
\dset{anneal} &$\thickapprox$ & $\thickapprox$ & $\thickapprox$ & $\thickapprox$ & $\thickapprox$ & $\thickapprox$ & $\thickapprox$ & $\thickapprox$  \\  
\dset{solar} &$\bigcirc$ & $\thickapprox$ & $\bigcirc$ & $\thickapprox$ & $\thickapprox$ & $\thickapprox$ & $\thickapprox$ & $\thickapprox$  \\  
\dset{splice} &$\thickapprox$ & $\bigcirc$ & $\thickapprox$ & $\bigcirc$ & $\thickapprox$ & $\bigcirc$ & $\thickapprox$ & $\thickapprox$  \\  
\dset{ecoli} &$\thickapprox$ & $\thickapprox$ & $\thickapprox$ & $\thickapprox$ & $\thickapprox$ & $\thickapprox$ & $\thickapprox$ & $\thickapprox$  \\  
\dset{nursery} &$\thickapprox$ & $\bigcirc$ & $\thickapprox$ & $\thickapprox$ & $\bigcirc$ & $\thickapprox$ & $\thickapprox$ & $\thickapprox$  \\  
\dset{soybean} &$\thickapprox$ & $\thickapprox$ & $\bigcirc$ & $\thickapprox$ & $\bigcirc$ & $\thickapprox$ & $\thickapprox$ & $\thickapprox$  \\  
\dset{arrhythmia} &$\bigcirc$ & $\thickapprox$ & $\bigcirc$ & $\thickapprox$ & $\thickapprox$ & $\thickapprox$ & $\thickapprox$ & $\thickapprox$  \\  
\dset{optdigits} &$\thickapprox$ & $\thickapprox$ & $\thickapprox$ & $\thickapprox$ & $\thickapprox$ & $\thickapprox$ & $\thickapprox$ & $\thickapprox$  \\  
\dset{mfeat} &$\thickapprox$ & $\thickapprox$ & $\thickapprox$ & $\thickapprox$ & $\bigcirc$ & $\thickapprox$ & $\thickapprox$ & $\thickapprox$  \\  
\dset{pendigit} &$\thickapprox$ & $\bigcirc$ & $\thickapprox$ & $\thickapprox$ & $\thickapprox$ & $\thickapprox$ & $\thickapprox$ \\
  \end{tabular}
  }
  \begin{tabular}{c@{$\colon$}l}
  $\bigcirc$ & soft cost-sensitive algorithms significantly better\\ 
  $\times$ & soft cost-sensitive  algorithms significantly worse\\ 
  $\thickapprox$ & otherwise
  \end{tabular}

    \end{center}
}
\end{table}

\begin{table}[h]
  \caption{comparison on the test errors between the hard cost-sensitive classification algorithms and their soft sibling
    using a pairwise one-tailed $t$-test of significance level $0.05$ in consistent cost}
  \vskip 0.1in
  \label{tbl:pt0111_test}
  {
    \begin{center}
   \scalebox{0.85}{
  \begin{tabular}{
      c||c|c|c|c
    }
    data set &  
	{OSR} &
	{CSOVO} &
	{CSFT}&
	{CSZL} \\
    \hline
    \dset{iris} &$\thickapprox$ & $\thickapprox$ & $\thickapprox$ & $\thickapprox$  \\  
\dset{wine} &$\thickapprox$ & $\thickapprox$ & $\thickapprox$ & $\thickapprox$  \\  
\dset{glass} &$\thickapprox$ & $\thickapprox$ & $\thickapprox$ & $\thickapprox$  \\  
\dset{vehicle} &$\thickapprox$ & $\thickapprox$ & $\thickapprox$ & $\thickapprox$  \\  
\dset{vowel} &$\bigcirc$ & $\bigcirc$ & $\bigcirc$ & $\thickapprox$  \\  
\dset{segment} &$\bigcirc$ & $\thickapprox$ & $\thickapprox$ & $\thickapprox$  \\  
\dset{dna} &$\bigcirc$ & $\thickapprox$ & $\thickapprox$ & $\thickapprox$  \\  
\dset{satimage} &$\thickapprox$ & $\thickapprox$ & $\thickapprox$ & $\thickapprox$  \\  
\dset{usps} &$\bigcirc$ & $\thickapprox$ & $\bigcirc$ & $\thickapprox$  \\  
\dset{zoo} &$\thickapprox$ & $\thickapprox$ & $\thickapprox$ & $\thickapprox$  \\  
\dset{yeast} &$\thickapprox$ & $\thickapprox$ & $\thickapprox$ & $\thickapprox$  \\  
\dset{pageblock} &$\thickapprox$ & $\thickapprox$ & $\thickapprox$ & $\thickapprox$  \\  
\dset{anneal} &$\bigcirc$ & $\thickapprox$ & $\thickapprox$ & $\thickapprox$  \\  
\dset{solar} &$\thickapprox$ & $\thickapprox$ & $\thickapprox$ & $\thickapprox$  \\  
\dset{splice} &$\bigcirc$ & $\bigcirc$ & $\bigcirc$ & $\thickapprox$  \\  
\dset{ecoli} &$\thickapprox$ & $\thickapprox$ & $\thickapprox$ & $\thickapprox$  \\  
\dset{nursery} &$\bigcirc$ & $\bigcirc$ & $\bigcirc$ & $\thickapprox$  \\  
\dset{soybean} &$\thickapprox$ & $\thickapprox$ & $\bigcirc$ & $\thickapprox$  \\  
\dset{arrhythmia} &$\thickapprox$ & $\thickapprox$ & $\thickapprox$ & $\thickapprox$  \\  
\dset{optdigits} &$\bigcirc$ & $\bigcirc$ & $\thickapprox$ & $\thickapprox$  \\  
\dset{mfeat} &$\bigcirc$ & $\bigcirc$ & $\thickapprox$ & $\thickapprox$  \\  
\dset{pendigit} &$\bigcirc$ & $\bigcirc$ & $\bigcirc$ & $\thickapprox$  \\  
  \end{tabular}
  }
  \begin{tabular}{c@{$\colon$}l}
  $\bigcirc$ & soft cost-sensitive algorithms significantly better\\ 
  $\times$ & soft cost-sensitive  algorithms significantly worse\\ 
  $\thickapprox$ & otherwise
  \end{tabular}
    \end{center}
}
\end{table}

\subsection{Comparison on a Real-world Biomedical Task}
To test the validity of our proposed soft cost-sensitive classification methodology on true applications, we use two real-world data sets for our experiments. The first one is a biomedical task \cite{bibm2011}, and the other one to be introduced later is from KDDCup 1999~\cite{bay2000uci}. Both data sets go through similar splitting and scaling procedures, as we did for the benchmark data sets.

The biomedical task is on classifying the bacterial meningitis, which is a serious and often life-threatening form of the meningitis infection. The inputs are the spectra of bacterial pathogens extracted by the Surface Enhanced Raman Scattering (SERS) platform \cite{campion1998surface}. In this paper, we call the task \dset{SERS}, which contains 79 clinical samples of ten
meningitis-causing bacteria species collected in the National Taiwan University Hospital and 17
standard bacteria samples from American Type Culture Collection. The cost matrix of \dset{SERS} is shown in Table \ref{costmatrix}, which is specified by two human physicians who are specialized in infectious
diseases.

\begin{table}[h]
    \caption{cost matrix on \dset{SERS}}
\label{costmatrix}
    \centering  	
    \vspace*{\baselineskip}
    \scalebox{0.7}{
      \begin{tabular}{|c|c|c|c|c|c|c|c|c|c|c|} 
        \hline   
        \backslashbox{real class}{classify to} & Ab & Ecoli &HI &KP&LM&Nm&Psa&Spn&Sa&GBS   \\[1ex] \hline
        Ab & 0 & 1 & {10}&7 & 9& 9 &5 &8&9&1 \\ \hline
        Ecoli & 3 & 0 & {10}&8 & 10& 10 &5 & {10} &{10}&2 \\ \hline
        HI & {10} & {10} & 0&3& 2& 2 &{10} &1&2&{10}\\ \hline
        KP & 7 & 7 & 3&0 & 4& 4 &6 &3&3&8 \\ \hline
        LM & 8 & 8 & 2&4 & 0& 5 &8 &2&1&8 \\ \hline
        Nm & 3 & {10} & 9&8 & 6& 0 &8 &3&6&7 \\ \hline
        Psa & 7 & 8 & {10}&9 & 9& 7 &0 &8&9&5\\ \hline
        Spn & 6 & {10} & 7&7 & 4& 4 &9 &0&4&7 \\ \hline
        Sa & 7 & {10} & 6&5 & 1& 3 &9 &2&0&7 \\ \hline
        Gbs & 2 & 5 & {10}&9 & 8& 6 &5 &6&8&0 \\ \hline
      \end{tabular} \label{tb:h1n1m}
    }
  \end{table} 

The results are shown in Table \ref{ff}. Among the eleven algorithms, soft-CSOVO gets the lowest cost. If we compare the other eight algorithms with soft-CSOVO using a pairwise one-tailed $t$-test of significance level $0.05$, we see that soft-CSOVO is significantly better than all other algorithms. The
results confirm the usefulness of soft cost-sensitive classification
for this real-world task.

\dset{SERS} is an interesting data set in which regular classification algorithms like OVO/ZL or FT can perform better than their hard cost-sensitive classification siblings like CSOVO or CSFT or CSZL. Given the small number of examples in \dset{SERS}, the phenomenon can be attributed to overfitting with respect to the cost---i.e. over-using the cost information. Soft cost-sensitive classification provides a balanced alternative between over-using (hard) or not using (regular) the cost. The balancing can lead to significantly lower test cost, as demonstrated by the promising performance of soft-CSOVO on this biomedical task. 

\begin{table}[h]
\caption{experiment results on \dset{SERS}, with $t$-test for cost}
\label{ff}
\begin{center}
\begin{tabular}{l|r|r|c}

      & error (\%) & cost ($\cdot 10^0$) & $t$-test \\
\hline
OVA   & $23.0\pm2.51$ & $1.056\pm0.097$ & $\bigcirc$ \\
\hline
OSR  & $ 27.6\pm2.27$ & $0.986\pm0.092 $ & $\bigcirc$ \\
\hline
soft-OSR  & $25.8\pm2.80$ & $ 1.024\pm0.095 $ & $\bigcirc$ \\
\hline
OVO/ZL & $23.2\pm2.55$ & $ 0.970\pm0.106$ & $\bigcirc$ \\
\hline
CSOVO & $ 27.4\pm1.53$ & $ 1.150\pm0.109$ & $\bigcirc$ \\
\hline
soft-CSOVO & $26.6\pm2.55 $ & $ 0.906\pm0.069$ & $*$\\
\hline
FT & $  23.0\pm2.51 $ & $ 0.986\pm0.092$ & $\bigcirc$ \\
\hline	
CSFT & $ 27.6\pm1.40$ & $1.118\pm0.090$ & $\bigcirc$ \\
\hline
soft-CSFT & $ 31.4\pm4.09$ & $ 1.054\pm0.040$ & $\bigcirc$ \\
\hline	
CSZL & $ 26.0\pm3.42$ & $1.030\pm0.110$ & $\bigcirc$ \\
\hline
soft-CSFT & $ 24.0\pm2.87$ & $ 0.990\pm0.105$ & $\bigcirc$ \\

\end{tabular}%
\end{center}
  \begin{tabular}{c@{$\colon$}l}
  $*$ & best entry of cost\\
  $\bigcirc$ & best entry significantly better in cost\\ 
  $\thickapprox$ & otherwise
  \end{tabular}

\end{table}

\subsection{Comparison on New Benchmark Tasks:\\ Emphasizing Cost }
Next, we explore the usefulness of the algorithms with a new benchmark. There are two situations when emphasizing different classes: The first situation is that one wants to indicate each class in the data set to be of different influence, which corresponds to scaling the \textit{rows} of the cost matrix as discussed in Section~\ref{section2}. The second situation is to avoid that the examples of  some classes to be wrongly predicted as some emphasized classes, which corresponds by scaling up some \textit{columns} of the cost matrix. As mentioned in Section~\ref{section2}, cost-sensitive classification is more sophisticated than re-weighting. In particular, it allows us to mark important classes by scaling up some \textit{columns} or some \textit{rows} of the cost matrix. In this benchmark task, we emphasize the \textit{columns} of the cost matrix by an emphasis \linebreak parameter $u$. 

%

\begin{figure*}[t]
  \centering
  \begin{tabular}{ccc}
  \subfigure[\dset{iris}]{
    \label{fig:subfig:iris} 
    \includegraphics[width=1.573in]{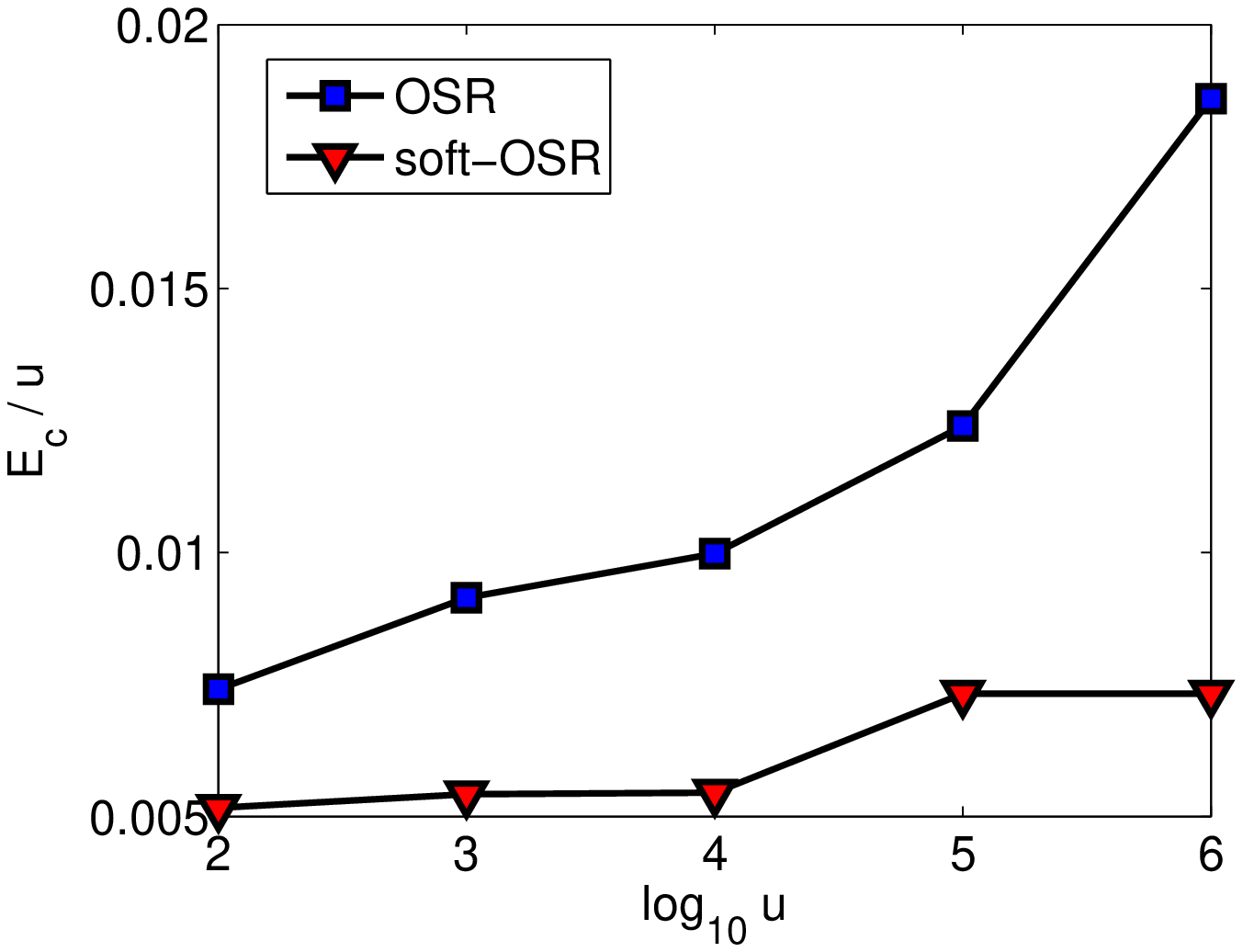}}
  &
  \subfigure[\dset{vehicle}]{
    \label{fig:subfig:vehicle} 
    \includegraphics[width=1.573in]{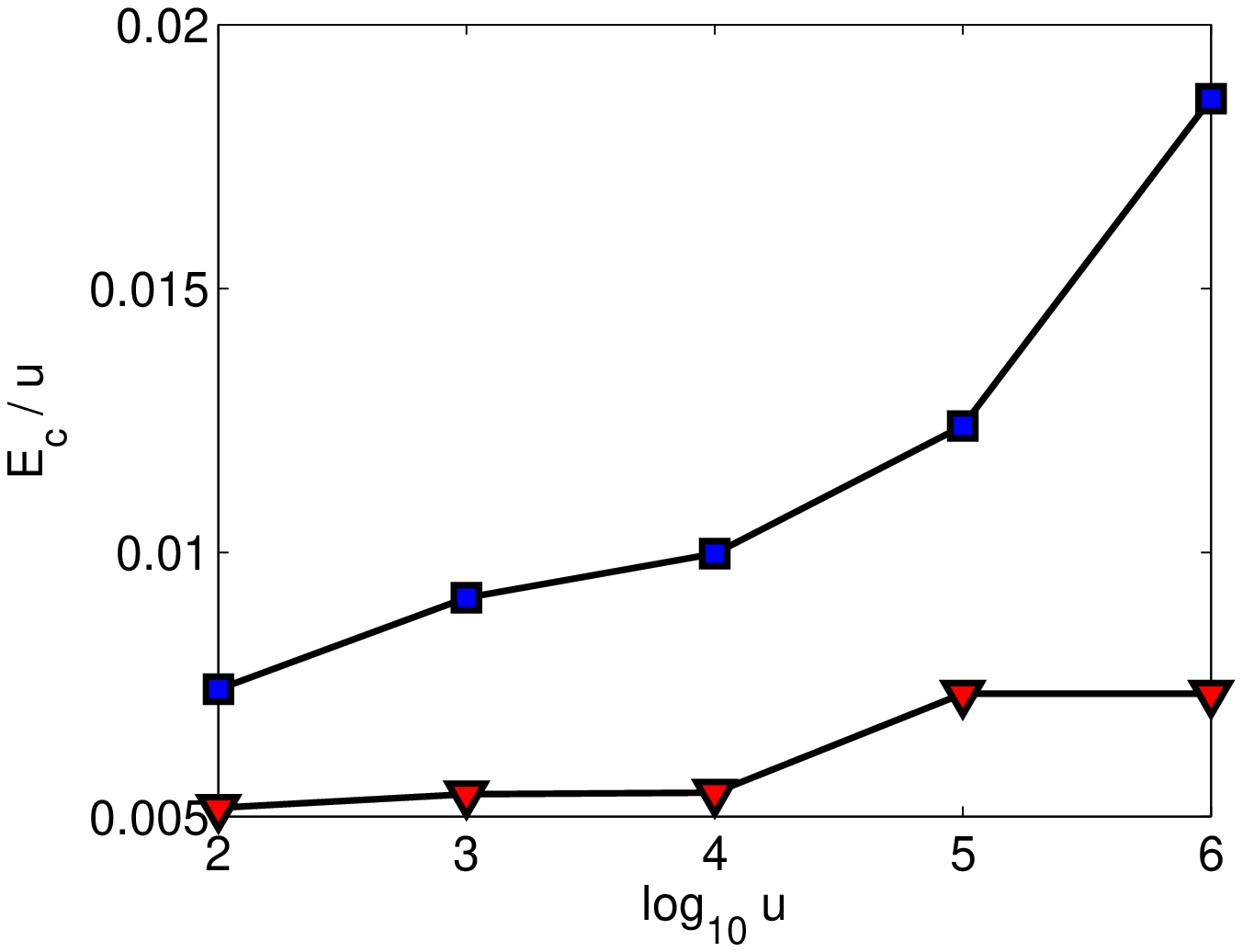}}
  &
  \subfigure[\dset{segment}]{
    \label{fig:subfig:segment} 
    \includegraphics[width=1.573in]{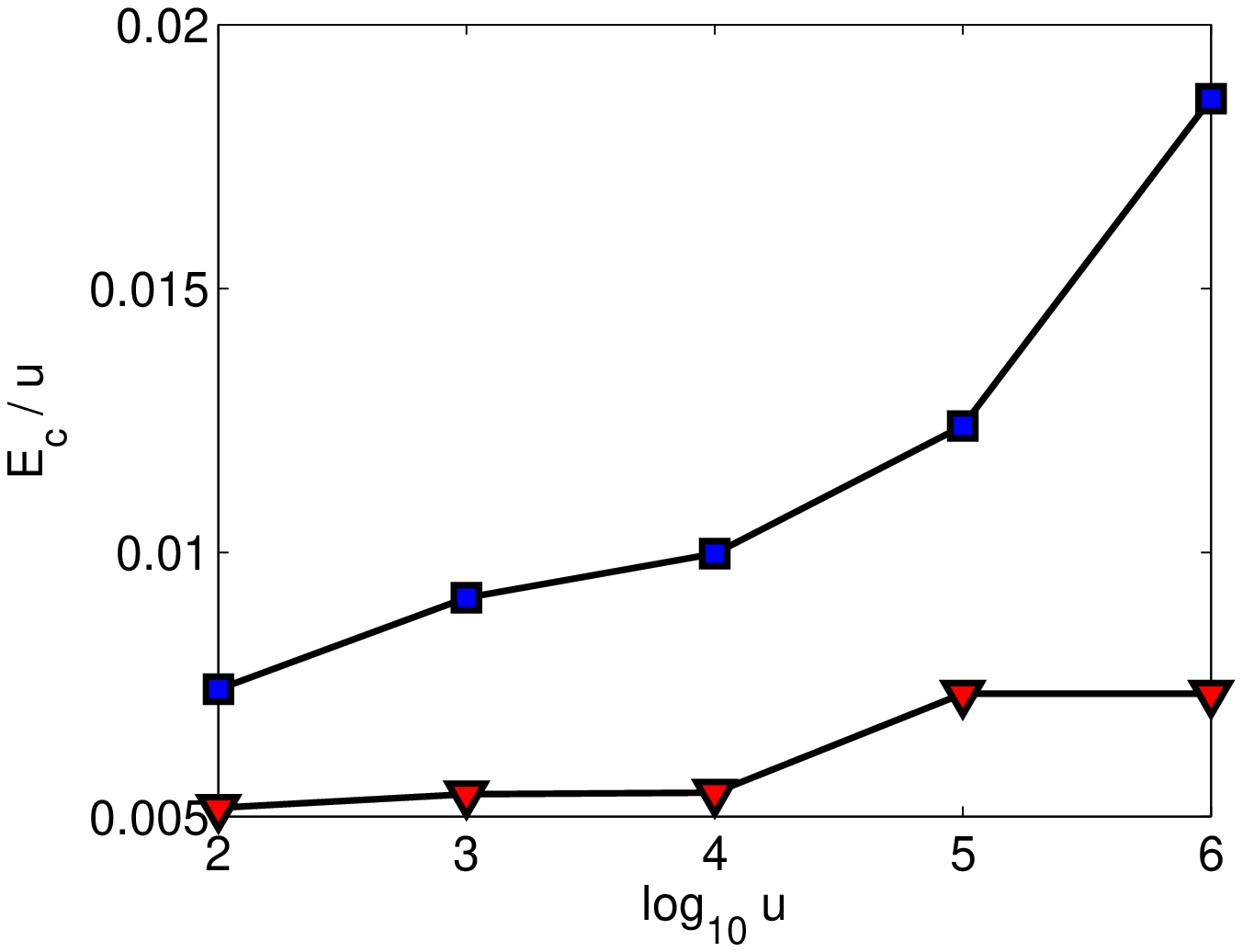}}        
  \end{tabular}
  \vskip -0.1in
  \caption{test $E_c / u$ of OSR and soft-OSR with the emphasizing cost for different emphasis parameter $u$}

  \label{fig:subfig2} 
\end{figure*}

We design the emphasizing cost to examine the stability of the algorithms when using large $u$. In this experiment, we vary the the emphasis  parameter $u$ between $\{10^2,10^3, \dots,10^6\}$. The results are shown in Figure~\ref{fig:subfig2}. Due to the page limits, we only report the results of OSR and soft-OSR on \dset{iris}, \dset{vehicle}, and \dset{segment}. The figures plot the scaled test cost $E_c / u$ on different values of $\log_{10} u$. From the three figures, we see that soft-OSR is better than OSR across all $\mbox{~}u$. When the emphasis is very high (like $10^6$), OSR can be conservative and ``paranoid.'' It avoids classifying any of the test examples as the emphasized class, which results in the worse performance. On the other hand,
the curves of soft-OSR remain mostly flat, which demonstrate that soft cost-sensitive classification is less sensitive (paranoid) to large cost components.
The results again justify the superiority of soft-OSR, a promising
representative of soft cost-sensitive classification, over its hard sibling.

%


%
%

\subsection{Comparison on New Benchmark Tasks:\\Unbalanced Classification}
The goal of this experiment is to examine the benchmark tasks with cost-sensitive and unbalanced data set. As discussed in Section \ref{section3}, weighted error rate is a more suitable basic criterion compared to error rate, and the corresponding methodology can be solved by using (\ref{softeq111}).

%
%
%
%

In this benchmark data set experiment, we  set $w_y = \frac{1}{|\{n:y_n\}|}.$ We further  scale every weight $w_y$ to   $\left[0,1\right]$ by dividing it with the largest component in weight. For the cost, we adapt the inconsistent benchmark cost mentioned in Section \ref{subsec1}. We choose ten unbalanced benchmark data sets, as shown in  Table \ref{tbl:data}. Then we compare three algorithms: OSR fed with the benchmark cost in Section \ref{subsec1}, weighted OVA with $w_y$ as weights, and soft OSR with the benchmark cost and weighted error. Table \ref{tbl:imba1} and Table \ref{tbl:imba2} show the cost and weighted error for those data sets. From Table \ref{tbl:imba1}, OSR achieve the lowest cost on most data set (except \dset{glass}); soft OSR is close to OSR in cost; weighted OVA falls behind. The results are similar to the findings in Section \ref{sub42}. On the other hand, from Table \ref{tbl:imba2}, weighted OVA reaches the lowest weighted error; OSR reaches the highest; soft-OSR is in between the two. The results justify that soft cost-sensitive classification can be used to achieve both low cost and low weighted error.

%




We further compare OSR with soft-OSR using another criterion: G-mean. G-mean is the geometric  mean accuracy of each class \cite{Alejo_improvingthe}. Higher G-mean reflects better performance for unbalanced classification tasks. The results are shown in Table \ref{tbl:gmean}. We see that soft-OSR out perform OSR in 8 out of 10 data sets. The results justify the usefulness of extending soft cost-sensitive classification with weighted error.

\begin{table}[h]
  \caption{unbalanced benchmark data sets}
  \vskip 0.1in
  \label{tbl:data}
  {
    \begin{center}
   \scalebox{0.8}{
  \begin{tabular}{
      c|c|c|c|c
    }
    data set &  
	 size &
	features  &
	class  &
	class distribution \\
    \hline
        \dset{pageblock}& 5473 & 10 & 5  & $ 4913 / 329 / 28 / 88 / 115  $ \\
\dset{wine} & 178 & 13 & 3  &$ 59 / 71 / 48 $\\
\dset{glass}& 214 & 9 & 6  &$ 70 / 76 / 17 / 13 / 9 / 29 $\\
\dset{dna}& 3186 & 180 & 3  &$ 767 / 765 / 1654  $ \\
\dset{satimage}& 6435 & 36  & 6 &$ 1533 / 703 / 1358 / 626 / 707 / 1508  $ \\
\dset{zoo}& 101 & 16 & 7  &$ 41 / 20 / 5 / 13 / 4 / 8 / 10  $ \\
\dset{yeast}& 1484 & 8 & 10  &$ 5 / 463 / 244 / 163 / 51 / 44 / 35 / 30 / 20 / 429  $ \\
\dset{anneal}& 898 & 84 & 5  & $ 684 / 40 / 8 / 67 / 99 $ \\
\dset{solar}& 1389 & 44 & 6  & $ 287 / 327 / 212 / 51 / 396 / 116  $ \\
\dset{splice}& 3190 & 287 & 3  &$ 767 / 768 / 1655  $ \\

  \end{tabular}
  }\\

\vskip \baselineskip
    \end{center}
}
\end{table}

\begin{table}[h]
  \caption{average test cost ($\cdot 10^{-3}$) on unbalanced benchmark data sets}
  \centering
  \label{tbl:imba1}
  {
  \scalebox{1}{
  \begin{tabular}{
      c|
            r@{$\pm$}l@{\hspace{7pt}}
        r@{$\pm$}l@{\hspace{7pt}}
                  r@{$\pm$}l@{\hspace{7pt}}
    }
  data set  & 
     \multicolumn{2}{c}{weighted OVA} &     
    \multicolumn{2}{c}{OSR}&
    \multicolumn{2}{c}{soft OSR}\\
    \hline
     \dset{pageblock} & $2.23$ & $0.11$ & $\mathbf{0.19}$ & $\mathbf{0.04}^*$ & $3.05$ & $0.49$ \\
\dset{wine} & $16.29$ & $3.68$ & $\mathbf{13.00}$ & $\mathbf{2.62}$ & $\mathbf{11.24}$ & $\mathbf{3.10}^*$ \\
\dset{glass} & $\mathbf{93.01}$ & $\mathbf{4.48}^*$ & $126.20$ & $9.84$ & $149.74$ & $10.39$ \\
\dset{dna} & $30.84$ & $1.01$ & $\mathbf{24.33}$ & $\mathbf{1.54}^*$ & $\mathbf{24.50}$ & $\mathbf{1.27}$ \\
\dset{satimage} & $53.46$ & $0.99$ & $\mathbf{35.18}$ & $\mathbf{2.14}^*$ & $40.07$ & $2.06$ \\
\dset{zoo} & $42.31$ & $5.29$ & $\mathbf{2.49}$ & $\mathbf{0.50}^*$ & $6.26$ & $1.81$ \\
\dset{yeast} & $13.60$ & $0.44$ & $\mathbf{0.58}$ & $\mathbf{0.07}^*$ & $36.66$ & $3.37$ \\
\dset{anneal} & $3.88$ & $0.66$ & $\mathbf{0.35}$ & $\mathbf{0.12}^*$ & $0.85$ & $0.23$ \\
\dset{solar} & $84.66$ & $2.17$ & $\mathbf{25.35}$ & $\mathbf{4.06}^*$ & $46.08$ & $6.53$ \\
\dset{splice} & $29.12$ & $1.20$ & $\mathbf{12.59}$ & $\mathbf{1.11}^*$ & $14.01$ & $0.84$ \\
  \end{tabular}
  }}\\
  
  (those with the lowest mean are marked with *; those within one standard error of the lowest one are in bold)    
\vskip -0.1in
\end{table}

  \begin{table}[h]
  \caption{average weighted error on unbalanced benchmark data sets}
  \centering
  \label{tbl:imba2}
  {
  \scalebox{1}{
  \begin{tabular}{
      c|
            r@{$\pm$}l@{\hspace{7pt}}
        r@{$\pm$}l@{\hspace{7pt}}
                  r@{$\pm$}l@{\hspace{7pt}}
    }
  data set  & 
     \multicolumn{2}{c}{weighted OVA} &     
    \multicolumn{2}{c}{OSR}&
    \multicolumn{2}{c}{soft OSR}\\
    \hline
     \dset{pageblock} & $\mathbf{3.06}$ & $\mathbf{0.15}^*$ & $21.40$ & $0.76$ & $20.49$ & $0.93$ \\
\dset{wine} & $\mathbf{0.73}$ & $\mathbf{0.17}^*$ & $1.48$ & $0.26$ & $\mathbf{0.79}$ & $\mathbf{0.16}$ \\
\dset{glass} & $\mathbf{5.02}$ & $\mathbf{0.24}^*$ & $5.81$ & $0.31$ & $5.73$ & $0.32$ \\
\dset{dna} & $\mathbf{24.58}$ & $\mathbf{0.81}^*$ & $42.29$ & $3.52$ & $30.18$ & $2.16$ \\
\dset{satimage} & $\mathbf{86.02}$ & $\mathbf{1.59}^*$ & $101.48$ & $3.35$ & $92.28$ & $2.46$ \\
\dset{zoo} & $\mathbf{1.10}$ & $\mathbf{0.14}^*$ & $2.28$ & $0.26$ & $\mathbf{1.20}$ & $\mathbf{0.20}$ \\
\dset{yeast} & $\mathbf{5.05}$ & $\mathbf{0.16}^*$ & $8.68$ & $0.18$ & $8.97$ & $0.29$ \\
\dset{anneal} & $\mathbf{0.87}$ & $\mathbf{0.15}^*$ & $2.71$ & $0.37$ & $1.63$ & $0.25$ \\
\dset{solar} & $\mathbf{29.46}$ & $\mathbf{0.75}^*$ & $35.51$ & $1.02$ & $34.71$ & $0.98$ \\
\dset{splice} & $\mathbf{23.23}$ & $\mathbf{0.96}^*$ & $49.84$ & $6.99$ & $28.07$ & $1.75$ \\
  \end{tabular}
  }}\\
  (those with the lowest mean are marked with *; those within one standard error of the lowest one are in bold)    
\vskip -0.1in
\end{table}

\begin{table}[h]
  \caption{average G-mean on unbalanced benchmark data sets}
  \centering
  \label{tbl:gmean}
  {
  \scalebox{1}{
  \begin{tabular}{
      c|
                  r@{$\pm$}l@{\hspace{7pt}}
        r@{$\pm$}l@{\hspace{7pt}}c 
    }
  data set  & 
    \multicolumn{2}{c}{soft OSR} &     
    \multicolumn{2}{c}{OSR}  &  
     $t$-test\\
    \hline
     \dset{pageblock} & $\mathbf{0.72}$ & $\mathbf{0.71}^*$ & $0.00$ & $0.00$ &  $\bigcirc$ \\
\dset{wine} & $\mathbf{94.08}$ & $\mathbf{1.23}^*$ & $87.25$ & $2.89$ &  $\bigcirc$ \\
\dset{glass} & $\mathbf{1.53}$ & $\mathbf{0.54}^*$ & $\mathbf{1.36}$ & $\mathbf{0.46}$ & $\thickapprox$ \\
\dset{dna} & $\mathbf{85.04}$ & $\mathbf{1.00}^*$ & $79.32$ & $1.59$ &  $\bigcirc$ \\
\dset{satimage} & $\mathbf{52.59}$ & $\mathbf{1.07}^*$ & $48.98$ & $1.46$ &  $\bigcirc$\\
\dset{zoo} & $\mathbf{9.17}$ & $\mathbf{4.93}^*$ & $0.00$ & $0.00$ &  $\bigcirc$\\
\dset{yeast} & $\mathbf{0.00}$ & $\mathbf{0.00}^*$ & $\mathbf{0.00}$ & $\mathbf{0.00}$ & $\thickapprox$ \\
\dset{anneal} & $\mathbf{39.88}$ & $\mathbf{7.22}^*$ & $19.59$ & $5.85$ & $\bigcirc$ \\
\dset{solar} & $\mathbf{0.69}$ & $\mathbf{0.25}^*$ & $0.45$ & $0.17$  & $\bigcirc$\\
\dset{splice} & $\mathbf{85.96}$ & $\mathbf{0.92}^*$ & $72.14$ & $4.81$ & $\bigcirc$ \\
  \end{tabular}
  
  }}\\
  \begin{tabular}{c@{$\colon$}l}
  $\bigcirc$ & soft cost-sensitive algorithms significantly better\\ 
  $\times$ & soft cost-sensitive  algorithms significantly worse\\ 
  $\thickapprox$ & otherwise
  \end{tabular}

  (those with the higest mean are marked with *; those within one standard error of the highest one are in bold)    
\vskip -0.1in
\end{table}

\subsection{Comparison on the KDD Cup 1999 Task: Cost-sensitive and unbalanced Classification}
The KDDCup 1999 data set (\dset{kdd99}) is another real-world cost-sensitive classification task~\cite{bay2000uci}. The task contains an intrusion detection problem
for distinguishing the ``good'' and
``bad'' connections.
Following the usual procedure in literature~\cite{NAb04}, we extract a random $40\%$ of the $10\%$-training set for our experiments. The test set accompanied is not used because of the known mismatch between training and test distributions~\cite{NAb04}. We take the given cost matrix in the competition for our experiments.%
\footnote{\url{http://www.kdd.org/kddcup/site/1999/files/awkscript.htm}} This data set is also highly unbalanced. In particular, the size of the majority class is over $8000$ times more than the size of the minority class. Therefore, we use the soft cost-sensitive classification and adapt weighted error rate as the basic criterion in this comparison. 

The results are listed in Table \ref{fff}. While the cost-sensitive classification algorithm OSR achieves the lowest test cost, other algorithms (soft, hard, or regular) all result in similar performance. The reason of the similar performance is because all the algorithms are of error rate less than $1\%$ and are thus of low weighted error and low cost. That is, the data set is easy to classify, and there is almost no room for improvements. 

 \begin{table}[h]
\caption{average test results on \dset{kdd99}, with $t$-test for cost}
\label{fff}
\begin{center}
\begin{tabular}{l|r|r|c}

      & error (\%) & cost ($\cdot 10^{-3}$) & $t$-test \\
\hline
OSR  & $ 0.11\pm0.003$ & $ 1.80\pm0.171$ & $*$ \\
\hline
soft-OSR  & $0.11\pm0.003$ & $ 1.89\pm0.163$ & $\thickapprox$ \\
\hline
CSOVO & $ 0.11\pm0.003$ & $  1.81\pm0.169$ & $\thickapprox$ \\
\hline
soft-CSOVO & $0.11\pm0.003 $ & $1.82\pm0.161$&$\thickapprox$\\
\hline	
CSFT & $0.11\pm0.003$ & $ 1.83\pm0.171$ & $\thickapprox$ \\
\hline
soft-CSFT & $ 0.11\pm0.003$ & $ 1.81\pm0.165 $ & $\thickapprox$ \\
\hline	
CSZL & $0.11\pm0.003$ & $ 1.83\pm0.170$ & $\thickapprox$ \\
\hline
soft-CSZL & $ 0.11\pm0.003$ & $ 1.81\pm0.162 $ & $\thickapprox$ \\

\end{tabular}

\end{center}
  \begin{tabular}{c@{$\colon$}l}
  $*$ & best entry of cost\\
  $\bigcirc$ & best entry significantly better in cost\\ 
  $\thickapprox$ & otherwise
  \end{tabular}
\end{table}

To further compare the performance of the algorithms, we consider
a more challenging version of the real-world task. The version is called
\dset{kdd99-balanced}, which adopted in our previous work \cite{soft2012}. The cost on \dset{kdd99-balanced} is scaled by the number of examples, which is generated by scaling down the $y$-th row of the cost matrix by the size of the $y$-th class.

The results on \dset{kdd99-balanced} are shown in  Table \ref{ffff}, and the $t$-test are listed in Table \ref{ffff1}. All algorithms share the similar cost except CSFT. However, soft cost-sensitive classification (with weighted error as the basic criterion) could reach the lower weighted error and the better G-mean significantly. The results again demonstrate the usefulness of soft cost-sensitive classification in reaching low cost \textit{and} low weighted error on this real-world task.


 \begin{table}[h]
\caption{average test results on \dset{kdd99-balanced}}
\label{ffff}
\begin{center}
\begin{tabular}{l|r|r|r}

      & weighted error & G-mean (\%)  &  cost ($\cdot 10^{-6}$) \\
\hline
OSR  & $ 9.10\pm0.56$ & $ 38.98\pm2.78 $ & $ 1.68\pm0.11 $ \\
\hline
soft-OSR  & $7.97\pm0.69$ & $  42.77\pm4.14 $ & $ 1.74\pm0.15 $ \\
\hline
CSOVO & $ 8.78\pm0.60$ & $  40.19\pm3.12$ & $ 1.63\pm0.10 $\\
\hline
soft-CSOVO & $8.28\pm0.55 $ & $ 41.63\pm3.35$&$ 1.68\pm0.13 $\\
\hline	
CSFT & $8.86\pm0.41$ & $ 36.98\pm1.51$ & $ 2.18\pm0.09 $ \\
\hline
soft-CSFT & $ 8.87\pm0.65$ & $  43.02\pm3.63$ & $ 1.66\pm0.12 $ \\
\hline	
CSZL & $9.84\pm0.59$ & $ 36.47\pm2.51$ & $ 1.85\pm0.06 $ \\
\hline
soft-CSZL & $ 9.02\pm0.73$ & $  41.39\pm4.01$ & $ 1.79\pm0.14 $ \\

\end{tabular}%

\end{center}
\end{table}

 \begin{table}[h]
\caption{$t$-test on \dset{kdd99-balanced} for weighted error, G-mean and cost}
\label{ffff1}
\begin{center}
\begin{tabular}{l|c|c|c}

      & weighted error & G-mean  &  cost  \\
\hline
OSR  & $ \bigcirc$ & $ \bigcirc  $ & $ \thickapprox $ \\
\hline
soft-OSR  & $*$ & $  \thickapprox $ & $ \thickapprox $  \\
\hline
CSOVO & $ \bigcirc $ & $  \thickapprox$ & $ * $\\
\hline
soft-CSOVO & $\thickapprox $ & $ \thickapprox $&$ \thickapprox $\\
\hline	
CSFT & $\bigcirc$ & $\bigcirc $ & $ \bigcirc$ \\
\hline
soft-CSFT & $ \bigcirc $ & $ *$ & $ \thickapprox$ \\
\hline	
CSZL & $\bigcirc$ & $\bigcirc $ & $ \bigcirc$ \\
\hline
soft-CSZL & $ \bigcirc $ & $ \thickapprox$ & $ \bigcirc$ \\

\end{tabular}%

\end{center}
  \begin{tabular}{c@{$\colon$}l}
  $*$ & best entry of the column\\
  $\bigcirc$ & best entry being significantly better \\ 
  $\thickapprox$ & otherwise
  \end{tabular}
\end{table}

%
%
%

\section{Conclusions}
\label{section5}

We have explored the trade-off between the cost and the error rate in cost-sensitive classification tasks, and have identified the practical needs
to reach both low cost and low error rate. 
Based on the trade-off, we have proposed
a simple and novel methodology between traditional regular classification
and modern cost-sensitive classification.
The proposed methodology, soft cost-sensitive classification,
takes both the cost and the error (or the weighted error) into account by a multicriteria optimization problem. 
By using the weighted sum approach to solving the optimization problem,
the proposed methodology allows immediate improvements of existing
cost-sensitive classification algorithms in terms of similar or sometimes lower costs, and of lower errors.
The significant improvements have been observed on a broad range of benchmark and real-world tasks in our extensive experimental study.



Our work reveals a new insight for cost-sensitive classification in machine learning and data mining: Feeding in the exact cost information for the machines to learn may not be the best approach, much like how fitting the provided data faithfully without regularization may lead to overfitting. Our work takes
the error rates to ``regularize'' the cost information and leads
to better performance. Another interesting direction for future research is to consider other types of regularization on the cost information.


%

\bibliographystyle{ACM-Reference-Format-Journals}
\bibliography{sigproc}

\received{}{}{}



\end{document}